\documentclass[iicol, pdflatex, sn-mathphys-num]{sn-jnl}

\usepackage{graphicx}%
\usepackage{multirow}%
\usepackage{amsmath,amssymb,amsfonts}%
\usepackage{amsthm}%
\usepackage{mathrsfs}%
\usepackage[title]{appendix}%
\usepackage{xcolor}%
\usepackage{textcomp}%
\usepackage{manyfoot}%
\usepackage{booktabs}%
\usepackage{listings}%
\usepackage{titlesec}

\usepackage{tikz}
\usetikzlibrary{arrows.meta,positioning,fit,calc,shapes.multipart}

\usepackage{subcaption}
\usepackage[ruled,linesnumbered]{algorithm2e}
\usepackage{comment}

\titleformat{\section}{\normalfont\fontsize{12}{12}\bfseries}{\thesection}{1em}{}
\titleformat{\subsection}{\normalfont\fontsize{11}{11}\bfseries}{\thesubsection}{1em}{}

\theoremstyle{thmstyleone}%
\theoremstyle{thmstyletwo}%

\theoremstyle{thmstylethree}%

\begin{document}

\title[Article Title]{Robotic System for Chemical Experiment Automation with Dual Demonstration of End-effector and Jig Operations}


\author*[1]{\fnm{Hikaru}    \sur{Sasaki}}
\author[1]{\fnm{Naoto}     \sur{Komeno}}
\author[1]{\fnm{Takumi}    \sur{Hachimine}}
\author[1]{\fnm{Kei}       \sur{Takahashi}}
\author[2]{\fnm{Yu-ya}     \sur{Ohnishi}}
\author[3]{\fnm{Tetsunori} \sur{Sugawara}}
\author[2]{\fnm{Araki}     \sur{Wakiuchi}}
\author[4]{\fnm{Miho}      \sur{Hatanaka}}
\author[5,6]{\fnm{Tomoyuki}  \sur{Miyao}}
\author[6]{\fnm{Hiroharu}  \sur{Ajiro}}
\author[5,6]{\fnm{Mikiya}    \sur{Fujii}}
\author[1,5]{\fnm{Takamitsu} \sur{Matsubara}}

%
%

\affil[1]{\orgdiv{Division of Information, Science Graduate School of Science and Technology}, \orgname{Nara Institute of Science and Technology},
\orgaddress{\street{8916-5, Takayama-cho}, \city{Ikoma}, \state{Nara}, \country{Japan}, \postcode{630-0192}}}

\affil[2]{\orgname{Materials Informatics Initiative, RD technology and digital transformation center, JSR Corporation},
\orgaddress{\street{3-103-9, Tonomachi, Kawasaki-ku},\city{Kawasaki}, \state{Kanagawa}, \country{Japan}, \postcode{210-0821}}}

\affil[3]{\orgname{Fine Chemical Process Dept., JSR Corporation},
\orgaddress{\street{100, Kawajiri-cho}, \city{Yokkaichi}, \state{Mie}, \country{Japan}, \postcode{510-8552}}}

\affil[4]{\orgdiv{Department of Chemistry Faculty of Science and Technology}, \orgname{Keio University},
\orgaddress{\street{3-14-1, Hiyoshi, Kohoku-ku}, \city{Yokohama}, \state{Kanagawa}, \country{Japan}, \postcode{223-8552}}}

\affil[5]{\orgdiv{Data Science Center}, \orgname{Nara Institute of Science and Technology},
\orgaddress{\street{8916-5, Takayama-cho}, \city{Ikoma}, \state{Nara}, \country{Japan}, \postcode{630-0192}}}

\affil[6]{\orgdiv{Division of Materials Science, Graduate School of Science and Technology}, \orgname{Nara Institute of Science and Technology},
\orgaddress{\street{8916-5, Takayama-cho}, \city{Ikoma}, \state{Nara}, \country{Japan}, \postcode{630-0192}}}





\abstract{
While robotic automation has demonstrated remarkable performance, such as executing hundreds of experiments continuously over several days, designing synchronized motions between the robot and experimental jigs remains challenging, especially for flexible experimental automation.
This challenge stems from the fact that even minor changes in experimental conditions often require extensive reprogramming of both robot motions and jig control commands.
Previous systems lack the flexibility to accommodate frequent updates, limiting their practical utility in actual laboratories.
To update robotic automation systems flexibly by chemists, we propose a concept that enables the automation of experiments by utilizing dual demonstrations of robot motions and jig operations by chemists.
To verify this concept, we developed a chemical-experiment-automation system consisting of jigs to assist the robot in experiments, a motion-demonstration interface, a jig-control interface, and a mobile manipulator.
We validate the concept through polymer-synthesis experiments, focusing on critical liquid-handling tasks such as pipetting and dilution.
The experimental results indicate high reproducibility of the demonstrated motions and robust task-success rates.
This comprehensive concept not only simplifies the robot programming process for chemists but also provides a flexible and efficient solution to accommodate a wide range of experimental conditions, providing a practical framework for intuitive and adaptable robotic laboratory automation.
Our project page is available at: \url{https://sasakihikaru.github.io/Chemical-Experiment-Automation-with-Dual-Demonstration/}.
}




\maketitle
\section{Introduction}
Advances in materials science have accelerated the cycle of theoretical verification through chemical experiments, requiring chemists to efficiently conduct a variety of experiments. 
To address the growing demand for chemical experiments, robotic automation has garnered significant attention \cite{burger2020, coley2019, fakhruldeen2022, yoshikawa2023c, lunt2024, chen2024, zhu2024, lim2021, tom2024}.
Notably, a robotic automation system utilizing a mobile manipulator and a dedicated laboratory successfully automated 668 experiments over an 8-day continuous operation period \cite{burger2020}.
For the common chemical laboratory task of liquid pipetting, jigs have been proposed to enable pipetting operations with robotic grippers that have limited degrees of freedom \cite{yoshikawa2023}.
Such robotic systems enable long-term repetitive experiments through the jig development and programming of the robot.
Despite their strength in repetitive execution, these systems often exhibit limited flexibility, as adapting them to other experiments typically demands time-consuming and technically intensive reprogramming.
This challenge is particularly critical in laboratory settings, where procedures frequently vary depending on the material, jigs, or measurement process.

To mitigate the cost of robotic programming, motion-demonstration interfaces for a human demonstrator unfamiliar with robotics have been proposed.
These interfaces are designed to match the shape of robotic manipulator end-effectors, enabling data collection suitable for robotic execution through human demonstrations \cite{hamaya2020, song2020a, young2021, chi2024}.
The commonality in shape between motion-demonstration interfaces and robotic end-effectors enables robots to directly imitate human-demonstrated motions.
However, robotic automation systems using these interfaces are limited to tasks executable by a robot's two-fingered gripper, making them unsuitable for complex tasks such as diverse chemical experiments.

\begin{figure*}[t]
    \centering
    \begin{minipage}[b]{0.45\linewidth}
        \centering
        \includegraphics[width=0.95\linewidth]{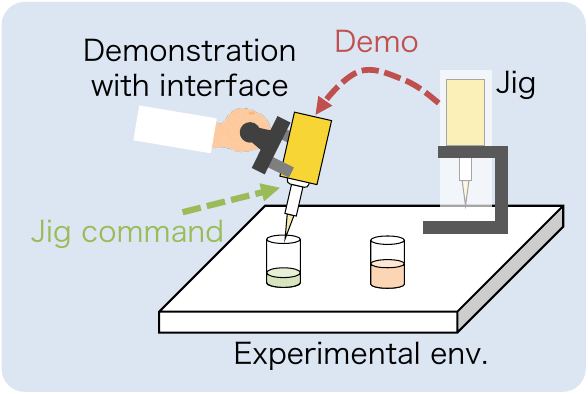}
        \subcaption{Demonstration}
        \label{fig:concept}
    \end{minipage}
    \begin{minipage}[b]{0.53\linewidth}
        \centering
        \includegraphics[width=0.95\linewidth]{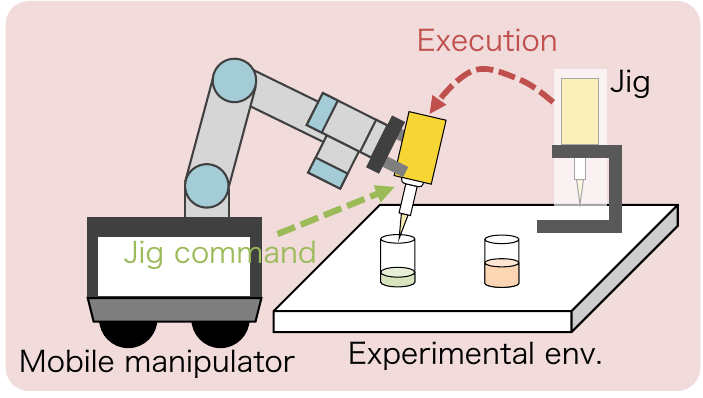}
        \subcaption{Execution}
        \label{fig:concept}
    \end{minipage}
    \caption{Proposed concept of dual demonstration of end-effector and jig operations for robotic-laboratory automation. (a) The system during a demonstration. A chemist conducts an experiment using jigs through a motion-demonstration interface. (b) The system is in execution. Mobile manipulator executes an experiment on a demonstration.}
    \label{fig:concept}
\end{figure*}

In recent laboratory automation systems, robots are increasingly expected to coordinate with multiple experimental jigs, such as pipettes.
However, as the number of such jigs grows, so does the complexity of programming coordinated motions between the robot and the jigs.
Previous approaches have either focused on manually programming these interactions or separating the robot's motion planning from jig operation, which makes it difficult to flexibly adapt procedures without expert intervention.
Prior work on robot teaching systems has mainly focused on acquiring effective motion data from human demonstrators, often assuming that device operations are preconfigured or managed separately.
We considered using the experimental jigs not only to assist the robot but also to assist the demonstration by chemists.
This would allow the coordinated motions involving both the robot and the experimental jigs to be demonstrated directly through the chemists' own task execution, without the need for manual programming.
By enabling them to directly demonstrate both robotic motions and jig operations without programming, the system can shorten experimental setup time, facilitate rapid adaptation to procedural changes, and improve the reproducibility of complex laboratory tasks. 
These features are particularly advantageous in environments where frequent adjustments and high-precision execution are essential.


In this paper, we propose a concept (Fig. \ref{fig:concept}) of dual-demonstration for achieving experimental automation through robotic-motion demonstration via a motion-demonstration interface and demonstration of jig utilization.
To validate this concept, we developed a chemical-experiment-automation system consisting of an experimental environment equipped with jigs, a motion-capture system comprising motion-demonstration and jig-control interfaces, and mobile manipulator reproducing demonstrated motions.
The experimental jigs, the motion-demonstration interface, a motion-capture system capturing interface movements, and the mobile manipulator are integrated via the robot operating system (ROS).
We focus on polymer synthesis in chemical experiments and developed experimental jigs, such as jigs designed for robotic-liquid manipulation, i.e., pipetting and bottle-holding.
Chemists use the motion-demonstration interface and a small joy controller to demonstrate robot motions and jig operations.
These dual demonstrations are recorded and subsequently used for robot and jig control.
Experimental verification involved polymer-synthesis tasks focusing on liquid sampling and dilution.
We evaluated our concept's capability to accurately replicate demonstrated actions and task-success rates.
The contributions of this study are summarized as follows.
\begin{itemize}
 \item Proposal of a concept using dual demonstration of robot motions and jig operations.
 \item A chemical-experiment-automation system consisting of jigs, motion-demonstration interface, and mobile manipulator through ROS, enabling robotic systems capable of dual demonstrations.
 \item In polymer-synthesis experiments, the mobile manipulator executed a dual demonstration with high precision, confirming sufficient performance for chemical-experiment automation.
\end{itemize}
Design files and control software are available on our project page: 

\noindent\textcolor{blue}{https://sasakihikaru.github.io/Chemical-Experiment-Automation-with-Dual-Demonstration/}.

\section{Related Works}
\subsection{Laboratory Automation}

Research on the automation of chemical experiments has rapidly advanced, with numerous approaches involving robots and automation systems being proposed.
A fully automated system using mobile robots demonstrated the capability to perform 688 experiments over 8 days \cite{burger2020}.
A platform integrating synthesis-prediction algorithms with robot-executable flow-synthesis devices has also been proposed, significantly improving experimental efficiency \cite{coley2019}.
The general-purpose robotic automation system ``ARChemist'' enables the integration of heterogeneous robots and chemical-experiment data, providing flexible system configurations through the ROS \cite{fakhruldeen2022}.
Approaches using large language models have shown new possibilities for chemical-experiment automation by converting scientists' natural language instructions into planning domain definition language (PDDL) and planning safe robot actions using PDDLstream \cite{yoshikawa2023c}.

For solid-state chemical experiments, modular-robot-integration approaches based on ARChemist technology have been proposed, enabling automated powder handling and transfer \cite{lunt2024}.
In inorganic material synthesis, robots executed 224 reactions involving 28 precursors and 27 elements to verify strategies for material selection \cite{chen2024}.
Optimization of catalyst synthesis derived from Martian meteorites has been achieved through methods combining robotic automation with machine learning, leading to approaches for materials discovery \cite{zhu2024}.
In organic-chemistry experiments, automation systems mimicking chemists' actions have been developed, incorporating optimized path-planning methods based on robot-posture data \cite{lim2021}.
Comprehensive reviews have also been conducted on autonomous laboratories in chemistry and materials sciences, clearly illustrating the field's overall progress and future directions \cite{tom2024}.

Despite advances in robotic technologies and proposals of general-purpose systems for diverse experimental automation, current methods still require task-specific robot-motion and jig-operation programming, posing challenges to the general applicability of automated systems.
In automated systems, as the number of jigs handled by the robot increases, the complexity of programming coordinated motions between the robot and jigs also increases.
This study proposes a dual demonstration system in which chemists can simultaneously operate a motion-demonstration interface and experimental jigs, enabling the execution of complex coordinated motions without the need for programming.

Our proposed concept leverages robotics-based demonstration systems for laboratory automation by simultaneously demonstrating robotic end-effector and jig operations.
We verified our developed chemical-experiment-automation system on the basis of this concept.

Table \ref{tab:comparison} summarizes the characteristics of representative laboratory automation systems in terms of task flexibility, execution precision, and required user expertise.
While ARChemist and self-driving laboratories offer strong automation capabilities, they typically require robotics or AI expertise to modify or extend.
Our proposed system provides high precision through demonstration-based execution and allows chemists who are not robotics experts to flexibly reconfigure tasks by directly demonstrating both robot and jig operations.

\begin{table*}[t]
    \centering
    \caption{Comparison of laboratory automation systems}
    \label{tab:comparison}
    \begin{tabular}{p{4.2cm} p{1.7cm} p{1.8cm} p{4.3cm}}
        \toprule
        \textbf{Automation system} & \textbf{Task flexibility} & \textbf{Execution precision} & \textbf{Required expertise} \\
        \midrule
        Mobile robot chemist \cite{burger2020}    & Medium        & High           & High (Robotics) \\
        Flow synthesis platforms \cite{coley2019} & Low           & High           & Medium (Domain-specific) \\
        ARChemist \cite{fakhruldeen2022}          & Medium        & High           & High (Robotics) \\
        Self-driving labs \cite{tom2024}          & High          & Medium         & High (AI/modeling) \\
        \textbf{Proposed system}                  & \textbf{High} & \textbf{High}  & \textbf{Low} \\
        \bottomrule
    \end{tabular}
\end{table*}

\subsection{Jigs for Laboratory Automation}
In laboratory automation, small-scale automation approaches focusing on specific tasks are gaining increasing attention. 
These approaches aim to support broader laboratory automation by enabling robots to perform well-defined subtasks efficiently. 
In particular, the design of experimental jigs plays a crucial role in realizing task-specific automation.

For liquid handling, several techniques have been developed that leverage specialized jigs.
For example, tactile-based posture estimation has been used with adaptive fingers to achieve precise well-plate insertion \cite{pai2023}. 
Uncalibrated vision systems and open-hardware jig designs have been successfully applied to pipette dispensing, enabling accurate liquid handling \cite{zhang2022, yoshikawa2023}. 
Syringe- and pipette-based liquid manipulation has also advanced through the use of custom-designed jigs \cite{wijnen2014, w.soh2023}. 
Similarly, 3D-printable jigs and linear-motion modules controlled by PCs have improved the precision of micro-pipetting tasks \cite{barthels2020, florian2020, dettinger2022}.

Solid-handling automation has also progressed with the use of jigs. 
For instance, soft jigs attached to robotic arms have enabled powder-grinding automation \cite{nakajima2022}, and reinforcement-learning-based control of jig mechanisms has facilitated delicate powder-handling tasks \cite{kadokawa2023}. 
Other approaches have used human-inspired controllers together with custom-designed jigs to automate narrow-domain tasks such as solid dispensing \cite{jiang2023a}.

These examples demonstrate that rather than attempting to automate entire experimental protocols, the careful design of jigs for individual tasks can substantially improve the precision, robustness, and flexibility of robotic execution in laboratory environments.

Previous studies on demonstration-based robot automation have primarily focused on how to efficiently acquire useful motion data from human demonstrators.
These approaches often emphasize learning or encoding end-effector trajectories suitable for execution by a robot.
In contrast, our work introduces the concept of jig-assisted demonstration into this paradigm.
By incorporating jigs that support precise and specialized manipulations, we propose a dual demonstration framework that enables chemists to demonstrate both motion and jig operations simultaneously.
This system extends the capabilities of conventional demonstration interfaces, allowing for the automation of expert-level experimental procedures without requiring manual programming.

\subsection{Robot Automation from Demonstration}
An increasing number of methods for robot automation based on human demonstrations have been proposed.
Interfaces equipped with cameras allow for intuitive motion demonstration, integrating general interfaces with robot-learning algorithms to enable policy learning from demonstrations in real environments \cite{young2021, chi2024}.
Research on soft robots has enabled the development of dedicated interfaces and visual-based imitation learning, capturing both successful and unsuccessful demonstration policies \cite{hamaya2020}.
High-precision pouring tasks have been automated using combinations of weight sensors attached to cups and trackers \cite{huang2021}.

Low-cost yet complex task applications have been explored through leader-follower teleoperation systems and demonstration interfaces with microphones, alongside grasping-position prediction methods using depth cameras \cite{fu2024, liu2024b, song2020a}.
Approaches for directly learning robotic policies from images have also advanced.
For example, visual teleoperation systems have been developed to recognize robot end-effector postures from human hand movements using a single camera, translating human hand motions into robotic actions \cite{handa2020a, qin2022}.
Other methods extract human hand trajectories and object movements from video data to learn robotic-control policies applied to specific tasks such as grasping \cite{jonnavittula2025}.
Simultaneous learning of image representation and robot actions or converting human-demonstration images to robot perspectives supports reinforcement learning and behavior prediction, requiring extensive data collection \cite{pari2022, li2021c, smith2020a}.
Efficient learning methods using inverse reinforcement learning and mutual information to assess deviations between demonstration data and robot observations have also been proposed \cite{cetin2020}.
Additional methods are used to plan category-level manipulation from single visual demonstrations using object computer-aided design and movement data, learn generalized manipulation policies from data collected by human-worn cameras, and remove humans and robots from demonstration videos \cite{wen2022b, bahl2022a}.


Recent advances in teleoperation and language-guided robot systems have explored complementary approaches to laboratory automation.
For instance, Yoshikawa et al. \cite{yoshikawa2023c} proposed a framework using large language models (LLMs) to convert natural language instructions into executable robot actions via PDDL-based planners.
While such systems offer high-level flexibility, they often require extensive domain knowledge modeling and task-specific configuration.
In contrast, our dual demonstration system enables low-level, direct demonstration of both robot motion and device control by chemists themselves, without the need for symbolic planning or expert-defined models.

These teleoperation or imitation learning based studies focused on learning highly generalized policies applicable to a variety of tasks and placed emphasis on optimal execution of demonstrated tasks, which required enormous amounts of time and computational resources.
Furthermore, because these methods target general operational tasks in general environments, they are not suitable for automating tasks that use specialized jigs, such as laboratory automation.
In contrast, this research proposes a novel automation method by focusing on the use of specialized jigs for chemical experiments.
Specifically, a dual demonstration that integrates the demonstration interface and the jig enables teaching of robot movements using a jig.

\begin{figure*}
    \centering
    \includegraphics[width=0.8\hsize]{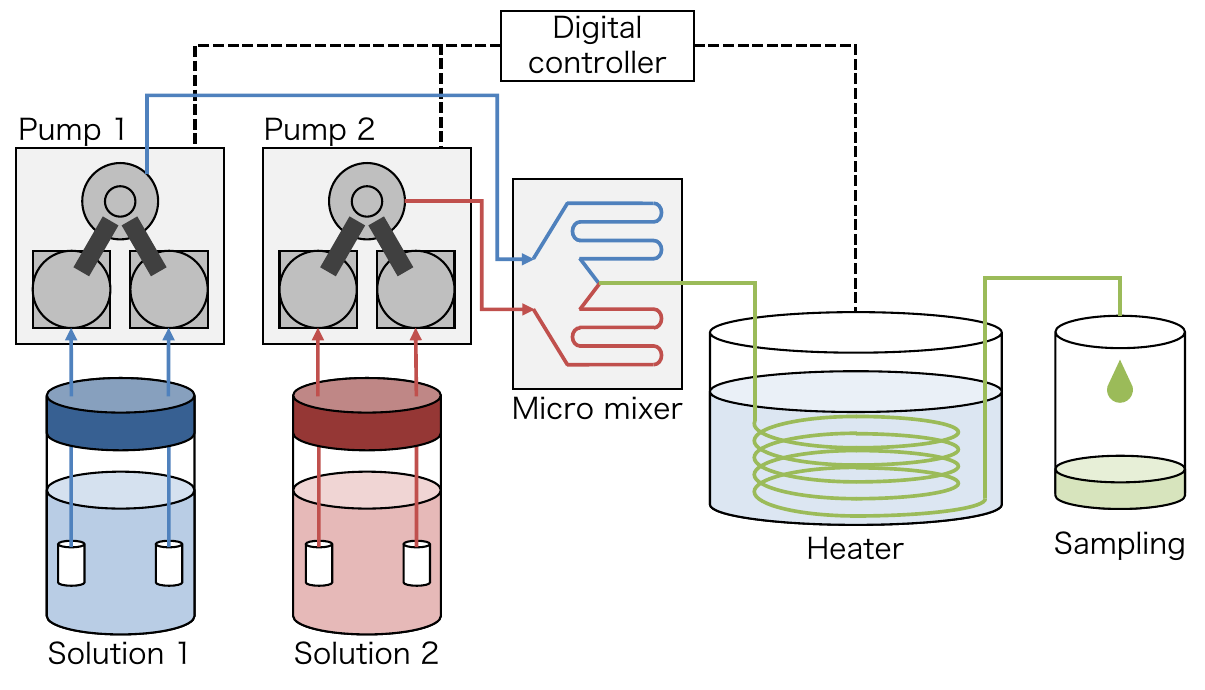}
    \caption{Flow polymerization system}
    \label{fig:flow_polymerization}
\end{figure*}
\begin{figure*}
    \centering
    \begin{minipage}[t]{0.49\linewidth}
        \centering
        \includegraphics[width=0.8\hsize]{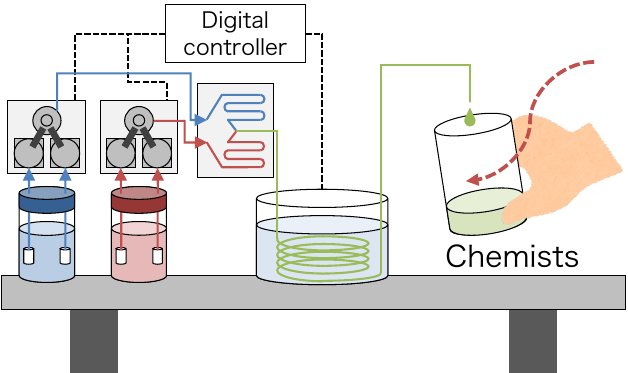}
        \subcaption{Polymer sampling}
    \end{minipage}
    \begin{minipage}[t]{0.49\linewidth}
        \centering
        \includegraphics[width=0.8\hsize]{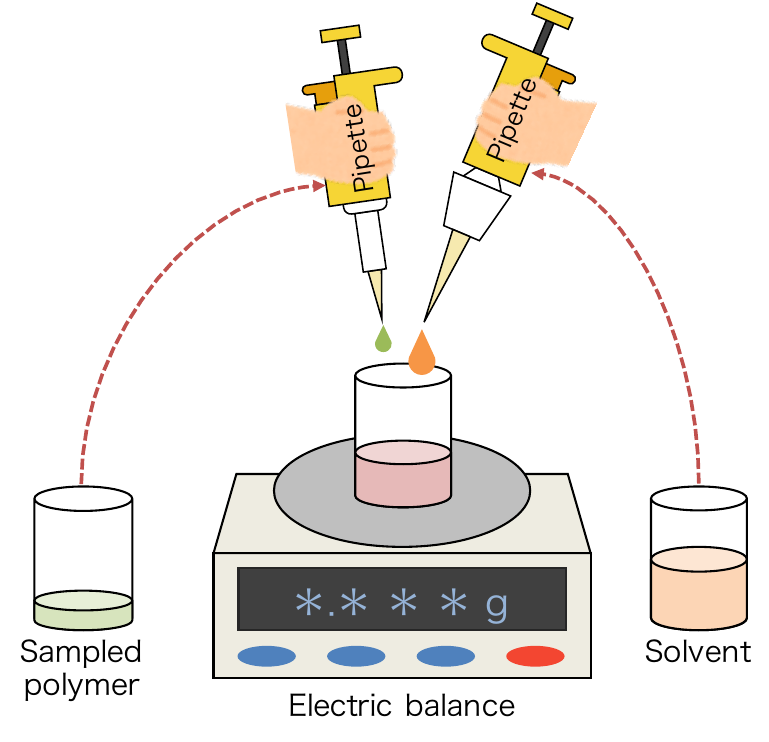}
        \subcaption{Sampled polymer dilution}
    \end{minipage}
    \vspace{-2mm}
    \caption{Manual tasks by chemists on flow copolymerization}
    \label{fig:flow_polymerization_tasks}
\end{figure*}

\section{Composition-regulation System by Flow Copolymerization}
We used a composition-regularization task involving a flow-copolymerization system \cite{wakiuchi2023} as an example of a chemical experiment.
This system efficiently produces polymers in various conditions by adjusting the mass fraction, residence time in the heat bath, and the pump-flow rate, as illustrated in Fig. \ref{fig:flow_polymerization}.
The pump-flow rate and reaction temperature in the heat bath can be digitally controlled via a computer.

In composition-regulation experiments using this system, chemists need to perform two tasks: sample the synthesized polymer and dilute the polymer as a preprocessing for analysis (Fig. \ref{fig:flow_polymerization_tasks}).
For a sample of the synthesized polymer, a bottle is used to sample the synthesized polymer coming out of the tube, and polymer dilution is used to dilute the sampled polymer into the analyzer.
To sample the polymer, chemists need to prepare a clean bottle and accurately transport it to the end of the tube with high precision.
To dilute the polymer, chemists need to operate a pipette and accurately measure the weight of both the polymer and dilution solution.
We focused on the sampling and dilution of polymer as manual tasks performed by chemists in polymer-synthesis experiments, and then applied our experiment-automation system to them.

\begin{figure*}[!t]
    \centering
    \begin{minipage}[b]{0.49\linewidth}
        \centering
        \includegraphics[width=\linewidth]{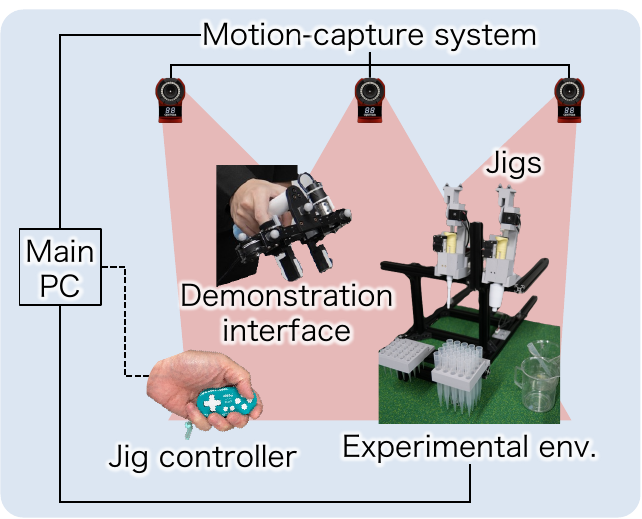}
        \subcaption{Demonstration}
        \label{fig:configuration:demonstration}
    \end{minipage} 
    \begin{minipage}[b]{0.49\linewidth}
        \centering
        \includegraphics[width=\linewidth]{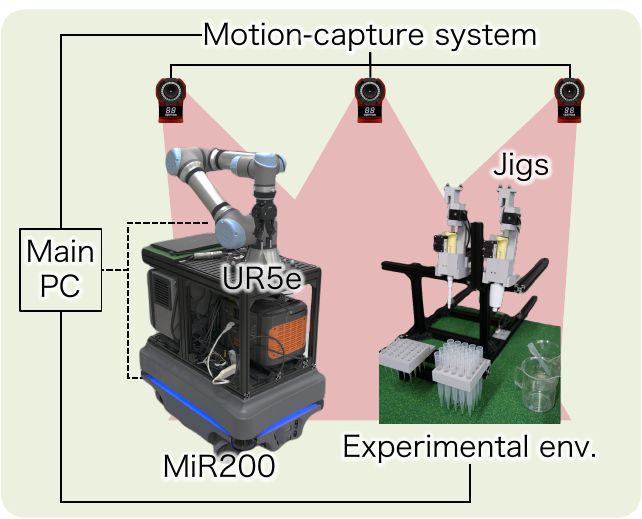}
        \subcaption{Execution}
        \label{fig:configuration:execution}
    \end{minipage}
    \caption{Overview of developed chemical-experiment-automation system. (a) Demonstration phase. Demonstrator conducts an experiment using a demonstration interface and jigs, and the interface's motions are collected using a motion-capture system. (b) Execution phase. Mobile manipulator executes an experiment based on a demonstration.}
    \label{fig:configuration}
\end{figure*}

\section{Proposed Concept of Dual Demonstration of End-effector and Jig Operations}
To overcome challenges in adapting robotic-automation systems to diverse chemical experiments, our proposed concept of dual demonstration integrates intuitive motion demonstrations with dedicated experimental jigs.
Traditional robotic systems often incur high reprogramming costs during task adjustments, limiting their flexibility and adaptability in real-world experimental environments.
While intuitive demonstration interfaces simplify robotic-motion demonstration, they typically lack the robustness and precision required for complex experimental procedures.
Conversely, jigs designed for specific tasks offer stability and precision but exhibit limited adaptability to new or modified tasks.

Our concept enables chemists to simultaneously demonstrate both robotic end-effector and jig operations.
This enables the collection of comprehensive motion data, enabling robots to effectively replicate the interactions between human demonstrators and tools.
By integrating a motion-demonstration interface that mimics the mobile manipulator's gripper structure with dedicated jigs controlled within a unified framework, our chemical-experiment-automation system combines the intuitive demonstration capabilities of demonstrations with the reliability and precision provided with experimental jigs.
This integrated approach significantly reduces programming complexity associated with changing automated chemical experiments, facilitating robotic automation across various chemical-experiment tasks.
Our system is expected to streamline robotic deployments in laboratory environments, promoting efficient and high-throughput chemical experimentation.

\begin{figure*}[t]
    \centering
    \includegraphics[width=0.95\linewidth]{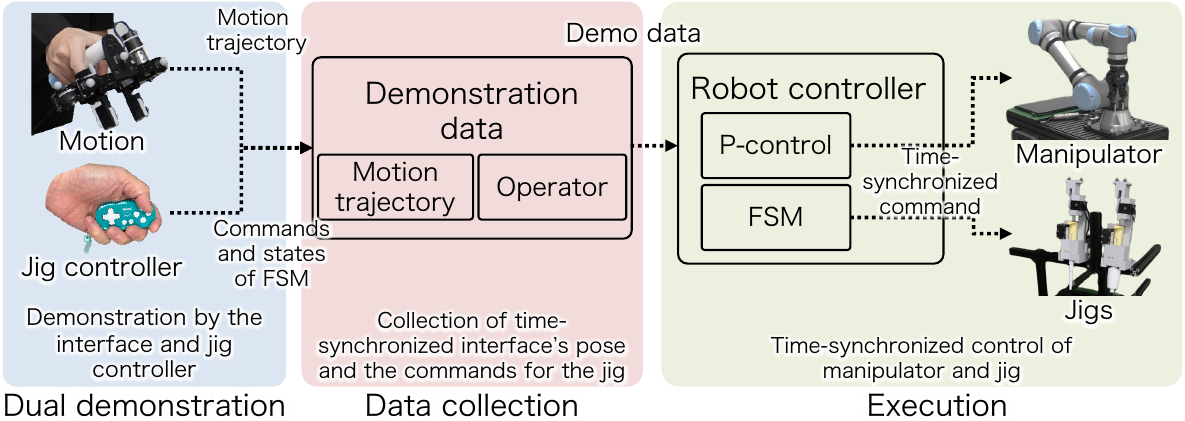}
    \vspace{1mm}
    \caption{
    Process flow of developed chemical-experiment-automation system.
    During the demonstration, the system synchronously records the pose data of the interface and the control commands of the jigs.
    These data are used to perform time-synchronized control of the manipulator and jigs, ensuring accurate and coordinated reproduction of the dual demonstration.
    }
    \label{fig:system:diagram}
\end{figure*}

\section{Chemical-experiment-automation System for Concept Verification}
To verify the proposed concept that robotic experimental automation can be achieved through demonstrations by chemists using a motion-demonstration interface and jig operation, we constructed our chemical-experiment-automation system for automating polymer-synthesis experiments.
The system overview and process flow are shown in Fig. \ref{fig:configuration} and Fig. \ref{fig:system:diagram}.

We developed an interface identical in shape to the robot's gripper for motion demonstration, jigs designed for pipetting and bottle handling, and a mobile manipulator.
Motion capture systems are utilized to collect demonstration data.
All these components are integrated via the ROS.

\begin{figure}[t]
    \centering
    \begin{minipage}[t]{0.49\linewidth}
        \centering
        \includegraphics[width=0.95\hsize]{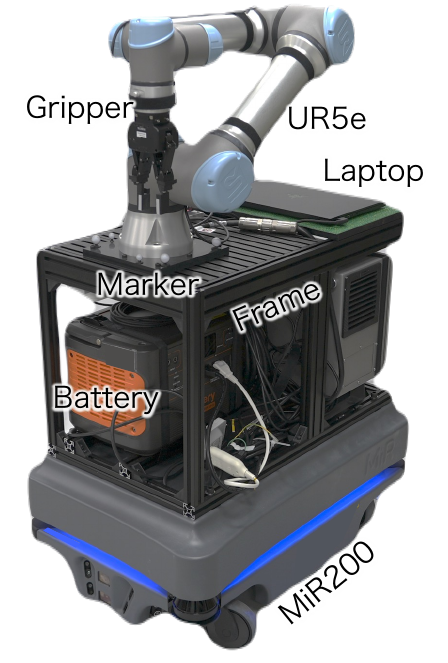}
        \subcaption{}
    \end{minipage}
    \begin{minipage}[t]{0.49\linewidth}
        \centering
        \includegraphics[width=0.95\hsize]{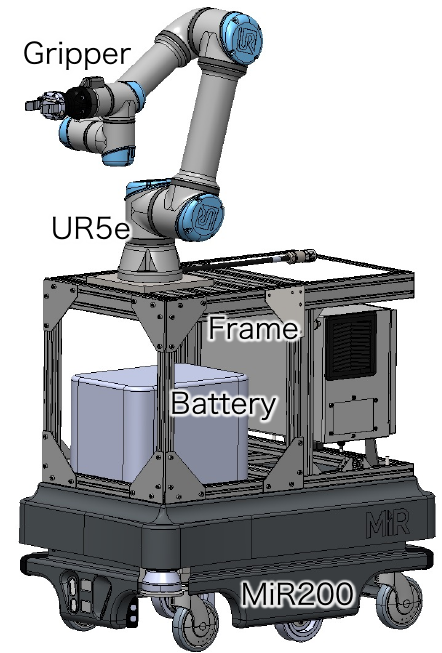}
        \subcaption{}
    \end{minipage}
    \caption{Mobile manipulator for our chemical-experiment-automation system. (a) Actual mobile manipulator. (b) 3D design of the mobile manipulator.}
    \label{fig:robot_system}
\end{figure}

\subsection{Mobile Manipulator}
Figure \ref{fig:robot_system} illustrates the mobile manipulator.
This mobile manipulator comprises a mobile robot and a manipulator.
Specifically, we used MiR200, a mobile robot from Mobile Industrial Robots, and UR5e, a manipulator from Universal Robots.
We use the ROBOTIS HAND RH-P12-RN by Robotis as the gripper.
The MiR200 mobile robot is a differential-drive mobile robot with a payload capacity of 200 kg.
The UR5e manipulator is a six-degree-of-freedom robotic arm with an operational radius of 850 mm and widely used in industrial applications due to its high repeatability of 0.03 mm.
The ROBOTIS gripper allows for fingertip-spacing adjustment between 0 to 106 mm.
The MiR200 mobile robot and UR5e manipulator are mounted together via an aluminum frame, positioning UR5e manipulator at table height.
A laptop onboard the robot controls the ROBOTIS gripper, while power is supplied to UR5e manipulator, laptop, and the ROBOTIS gripper with a Jackery Portable Power 1500 unit.
Motion-capture markers attached to the base of UR5e manipulator provide positional data essential for controlling the robot's end-effector.
The MiR200 mobile robot, UR5e manipulator, and the ROBOTIS gripper components are connected via Wi-Fi and integrated via the ROS.

The MiR200 mobile robot uses onboard LiDAR (Light Detection and Ranging) and cameras to map the room, enabling navigation to designated goal positions within the created map.
With our system, known positions for MiR200 mobile robot during robot replication of human actions are assumed, relying on navigation capabilities to reach these positions.
The UR5e manipulator uses URscript, a specialized script for UR robots, to implement end-effector velocity control.
ROS-based communication transmits velocity commands derived from demonstration data to UR5e manipulator.

\begin{figure*}[t]
    \centering
    \begin{minipage}[b]{0.38\linewidth}
        \begin{minipage}[b]{\linewidth}
            \centering
            \includegraphics[width=\hsize]{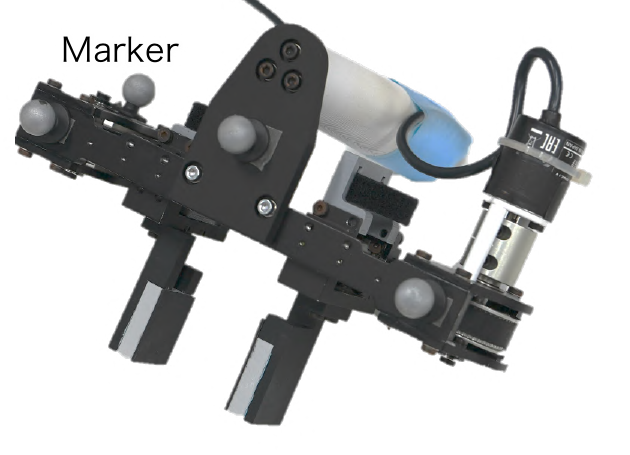}
            \subcaption{}
            \label{fig:teaching_device:view}
        \end{minipage} \\
        \begin{minipage}[b]{\linewidth}
            \centering
            \includegraphics[width=\hsize]{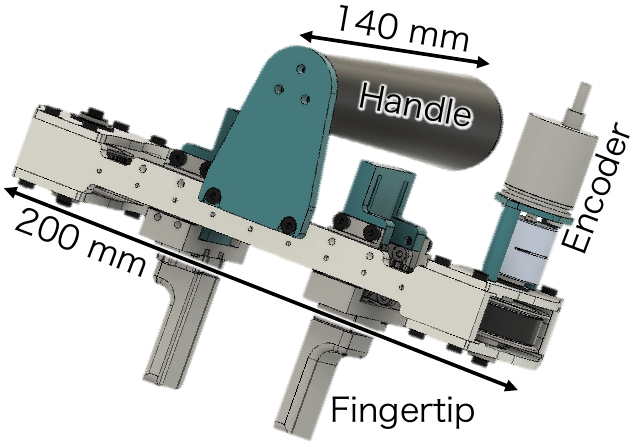}
            \subcaption{}
            \label{fig:teaching_device:design}
        \end{minipage} 
    \end{minipage} 
    \begin{minipage}[b]{0.32\linewidth}
        \centering
        \includegraphics[width=\hsize]{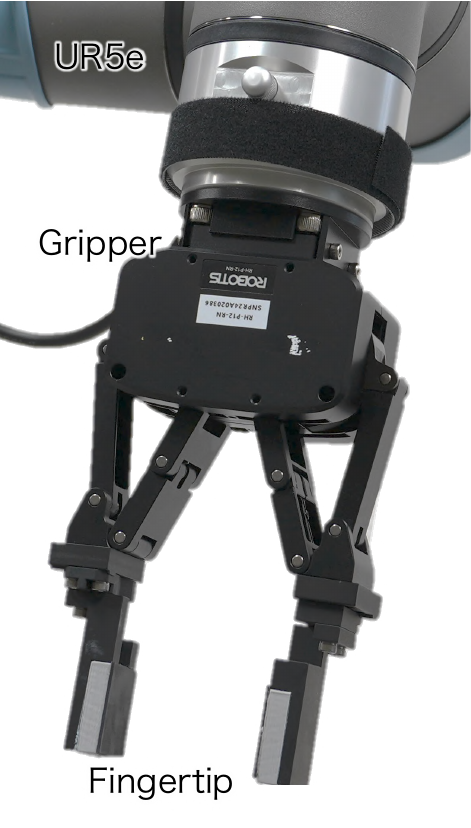}
        \subcaption{}
        \label{fig:teaching_device:robot_gripper}
    \end{minipage}
    \caption{Design of motion demonstration interface. (a) Overview of the motion demonstration interface. (b) Detailed design of the interface. Shape of fingertips is shared with the mobile manipulator's gripper, as shown in (c).} 
    \label{fig:teaching_device}
\end{figure*}

\subsection{Motion-demonstration Interface}
We developed the motion-demonstration interface to be structurally aligned with the ROBOTIS gripper used with our chemical-experiment-automation system.
Figure \ref{fig:teaching_device} shows this motion-demonstration interface alongside the mimicking the ROBOTIS gripper structure.
The interface is 200 mm wide with a 140-mm handle and can open its fingers up to 80 mm.
Since the ROBOTIS gripper can open up to 106 mm, it effectively replicates the finger movements demonstrated via the interface.
An Omron rotary encoder (e6b2-cwz3c) attached to the interface captures finger positions, and an Arduino Uno transmits these encoder readings.
The Arduino Uno is integrated with the ROS to relay finger-position data.
The interface mimics the structure of the ROBOTIS gripper, featuring two fingertips that match the ROBOTIS gripper's design, ensuring consistency in fingertip geometry between the robot and demonstration interface.
Users grip the handle and manipulate the interface with two fingers, performing the desired robotic tasks during demonstration.

For tracking the interface's position within the workspace, markers attached to the motion-demonstration interface are monitored with an OptiTrack Flex13 optical motion-capture camera at 120 (Hz).
The motion-capture system offers sub-millimeter accuracy with a typical positional error of less than $\pm$ 0.10 (mm) under well-calibrated conditions.
Given this high level of accuracy, we did not incorporate an explicit measurement noise model or uncertainty estimation.
For this work, this level of precision was sufficient to ensure reliable execution without the need for additional filtering or compensation mechanisms.
The motion data are processed via Motive software on a Windows PC integrated with the ROS, facilitating direct robotic execution of demonstrated chemical-experiment tasks.

To accurately reproduce human-demonstrated motions using the robot, the positions of the motion-demonstration interface and the robot's end-effector must be appropriately aligned.
The motion-capture system estimates the position of a rigid body based on the centroid of its attached markers.
To ensure that the measured position corresponds to the center between the fingertips of the motion-demonstration interface, we placed the marker cluster such that its geometric center aligns with the midpoint between the two fingers. 
This location approximates the tool center point (TCP) of the robot's gripper, allowing the recorded demonstrations to be directly transferred to robot execution without the need for additional calibration.

\subsection{Experimental Jigs}
To automate liquid-handling in experiments through motion demonstrations and robotic execution, we developed the experimental automation jigs to be compatible with two-fingered grippers.
These jigs establish consistent experimental environments for both human demonstrations with the interface and robotic executions.
To maintain consistency between demonstrations and executions, all experimental jigs are rigidly mounted to the laboratory environment. This physical mounting ensures that the jigs are placed in fixed, repeatable positions relative to the robot, thereby minimizing spatial misalignment and facilitating accurate reproduction of tool trajectories and orientations.
The experimental jigs are designed to support manipulators' motions that would otherwise require compliance control or explicit modeling of physical interactions, enabling stable and repeatable manipulation without the need for additional sensing or complex dynamical models.
These jigs were specifically designed for pipetting, bottle mounting, flow plumbing, and bottle holding.

We used ROS-compatible actuators comprising Robotis Dynamixel XM430-W350 servo motors controlled via U2D2 interfaces.
These servo motors use daisy-chain connections, enabling straightforward integration and easy device interchangeability tailored to specific experimental requirements.
Device components were manufactured using a FLASHFORGE Creator3 Pro 3D printer and PLA filament material.

We adopted finite-state machines (FSMs) for controlling the experimental jigs due to their simplicity and interpretability.
FSMs allow for a clear definition of discrete jigs states and deterministic transitions based on control inputs, which aligns well with the structured and repeatable nature of laboratory tasks.
The FSM states and state transitions of each jig are defined according to its function, facilitating experiment execution with robot and interface inputs.
During the demonstrations, chemists used an ``8Bitdo Zero 2'' wireless controller with their free hand to send device control commands, complementing the motion-demonstration interface.
Because each FSM transition is triggered only by explicit control commands and the tasks are executed in a quasi-static manner, the system does not require additional handling of nonlinear dynamics or time delays.


\begin{figure*}[!t]
    \centering
    \begin{minipage}[b]{0.35\linewidth}
        \centering
        \includegraphics[width=0.8\hsize]{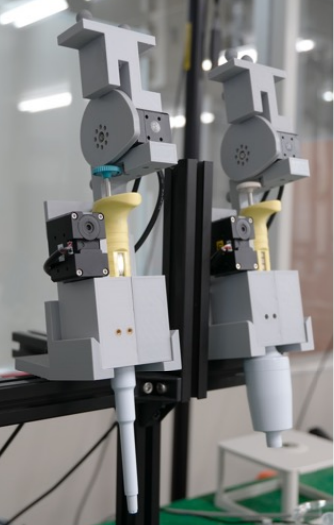}
        \subcaption{}
        \label{fig:jig:pipette:view}
    \end{minipage}
    \begin{minipage}[b]{0.22\linewidth}
        \centering
        \includegraphics[width=0.8\hsize]{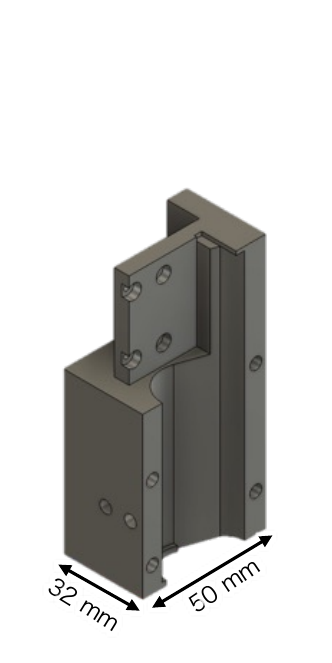}
        \subcaption{}
        \label{fig:jig:pipette:left}
    \end{minipage}
    \begin{minipage}[b]{0.28\linewidth}
        \centering
        \includegraphics[width=0.8\hsize]{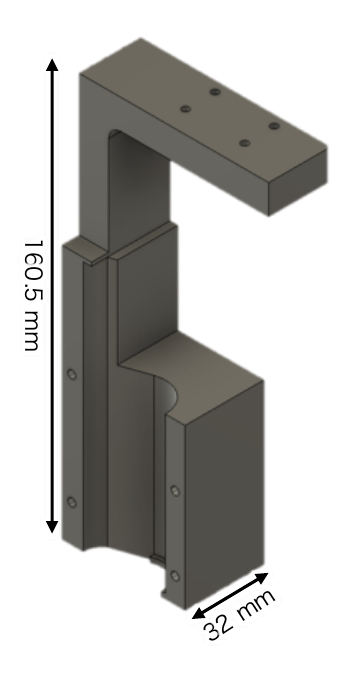}
        \subcaption{}
        \label{fig:jig:pipette:right}
    \end{minipage}\\
    \begin{minipage}[b]{0.32\linewidth}
        \centering
        \includegraphics[width=0.7\hsize]{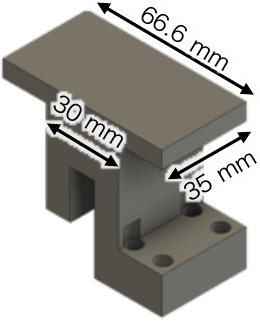}
        \subcaption{}
        \label{fig:jig:pipette:handle}
    \end{minipage}
    \begin{minipage}[b]{0.32\linewidth}
        \centering
        \includegraphics[width=0.8\hsize]{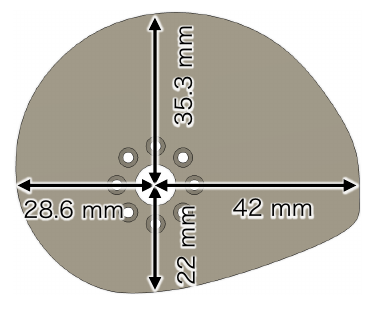}
        \subcaption{}
        \label{fig:jig:pipette:pusher}
    \end{minipage}
    \begin{minipage}[b]{0.32\linewidth}
        \centering
        \includegraphics[width=0.6\hsize]{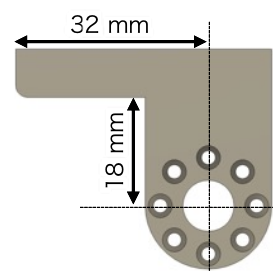}
        \subcaption{}
        \label{fig:jig:pipette:tip_pusher}
    \end{minipage} 
    \caption{Pipetting jig: (a) photo of entire jig, (b) left outer casing, (c) right outer casing, (d) handle, (e) plunger pusher, and (f) tip-ejector-button pusher}
    \label{fig:jig:pipette}
\end{figure*}

\subsubsection{Pipetting Jig}
To automate pipetting tasks using the mobile manipulator and motion-demonstration interface, we developed the pipetting jig shown in Fig. \ref{fig:jig:pipette}.
The jig incorporates two types of micropipettes: Nichipet EX Plus II 00-NPLO2-1000 for 100- to 1000-$\mu$L liquid handling and Nichipet EX Plus II 00-NPLO2-10000 for 1- to 10-mL liquid handling.
By integrating 3D-printed enclosures and actuators, the jig enables digital control of both the plunger pusher and tip-ejector buttons.

The pipetting jig consists of five 3D-printed components and two servo motors.
Figures \ref{fig:jig:pipette:left} and \ref{fig:jig:pipette:right} show the device's left and right outer casings, respectively.
Figure \ref{fig:jig:pipette:handle} displays the handle component, designed for gripping with the motion-demonstration interface or the mobile manipulator's gripper, and mounted on top of this jig.
Figures \ref{fig:jig:pipette:pusher} and \ref{fig:jig:pipette:tip_pusher} show the components that press the pipette's plunger and tip-ejector buttons, respectively.
These components are actuated via servo motors that control the rotation amount.
The plunger pusher features a non-circular cam shape, enabling the insertion depth to vary according to the rotation angle.

The pipetting jig uses an FSM for each servo motor to control the plunger and tip-ejection operations.
The FSM for plunger control defines three states: ``released'', ``pressed to 1st stop'', and ``pressed to 2nd stop''.
These states correspond to specific servo-rotation angles, and transitions between states trigger the motor to move to the designated angle.
Each time the FSM for plunger control receives a control command, it transitions between states in a loop: ``pushed to 1st stop'', ``released'', ``pushed to 2nd stop'', ``released''.

The FSM for tip ejection includes two states: ``button released'' and ``button pressed''.
Upon receiving a control command, the FSM transitions from ``button released'' to ``button pressed'', and after a 3-second delay, it automatically transitions back to the ``button released'' state.
This design enables accurate and consistent robotic pipetting operations in experimental tasks.

\begin{figure*}
    \centering
    \begin{minipage}[b]{0.4\linewidth}
        \centering
        \includegraphics[width=0.9\hsize]{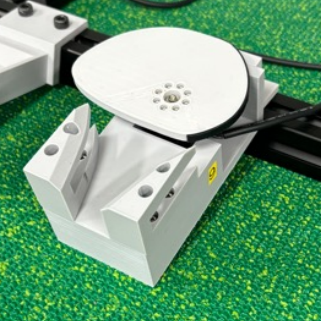}
        \subcaption{}
        \label{fig:jig:bottle_mounter:overview}
    \end{minipage}
    \begin{minipage}[b]{0.4\linewidth}
        \centering
        \includegraphics[width=0.9\hsize]{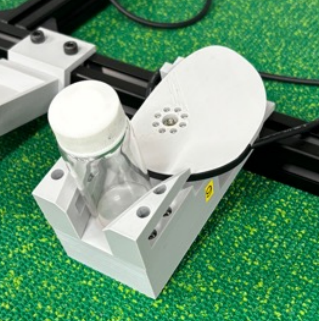}
        \subcaption{}
        \label{fig:jig:bottle_mounter:overview}
    \end{minipage}  \\
    \begin{minipage}[b]{0.4\linewidth}
        \centering
        \includegraphics[width=0.9\hsize]{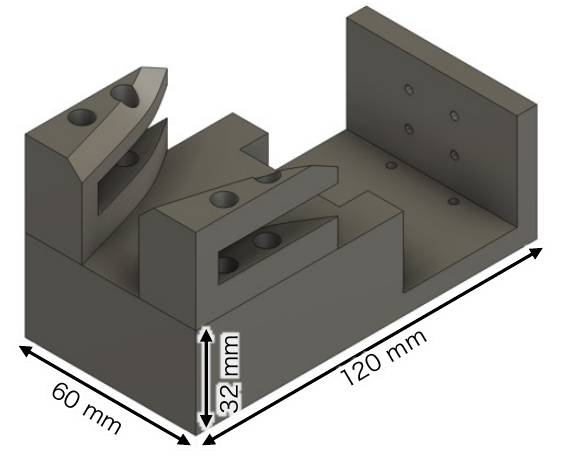}
        \subcaption{}
        \label{fig:jig:bottle_mounter:base}
    \end{minipage}
    \begin{minipage}[b]{0.4\linewidth}
        \centering
        \includegraphics[width=0.9\hsize]{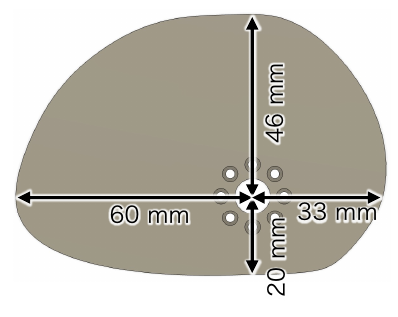}
        \subcaption{}
        \label{fig:jig:bottle_mounter:pusher}
    \end{minipage}
    \caption{Bottle-mounting jig: (a) photo of entire jig. It consists of base, pusher, and Dynamixel servo motor. (b) Jib fixes bottle by rotating plate. (c) and (d) Design information for two parts.}
    \label{fig:jig:bottle_mounter}
\end{figure*}

\subsubsection{Bottle-mountering Jig}
The bottle-mounting is designed to hold bottles securely, enabling the motion-demonstration interface or mobile manipulator to open and close bottle caps using a single hand.
This jig is shown in Fig. \ref{fig:jig:bottle_mounter}.
It consists of a base with two gripping claws (Fig. \ref{fig:jig:bottle_mounter:base}), non-circular rotating plate that presses bottles against the claws (Fig. \ref{fig:jig:bottle_mounter:pusher}), and servo motor that rotates the plate.

When the plate is rotated with the servo motor, it pushes the bottle against the two claws of the base, securing it in place.
This setup supports the ROBOTIS gripper or interface in executing cap -opening and -closing operations.
The servo motor operates in current-based position-control mode within a predefined load range, enabling the rotating plate to continue rotating until it contacts the bottle.
This mechanism accommodates bottles of various sizes by enabling adaptive fixation.

The FSM controlling the bottle mounter defines two states: ``locked'' and ``unlocked''.
When a control command is received, the FSM of this jig toggles between these two states, adjusting the position of the rotating plate accordingly to secure or release the bottle.

\begin{figure*}[t]
    \centering
    \begin{minipage}[b]{0.6\linewidth}
        \centering
        \includegraphics[width=\hsize]{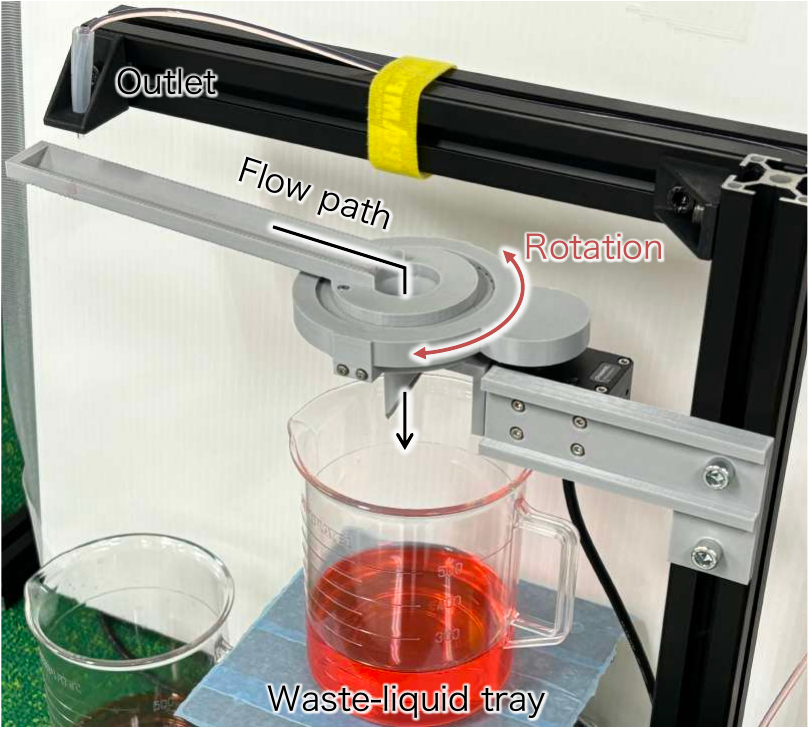}
        \subcaption{Overview}
        \label{fig:jig:plumbing:overview}
    \end{minipage} 
    \begin{minipage}[b]{0.3\linewidth}
        \begin{minipage}[b]{\linewidth}
            \centering
            \includegraphics[width=\hsize]{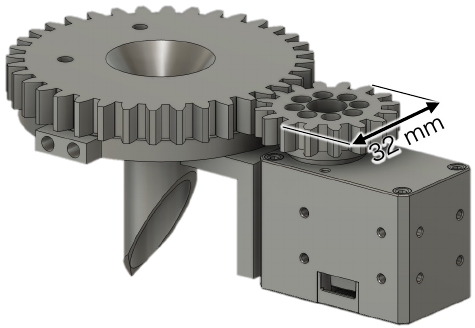}
            \subcaption{}
            \label{fig:jig:plumbing:rotor}
        \end{minipage} 
        \begin{minipage}[b]{\linewidth}
            \centering
            \includegraphics[width=\hsize]{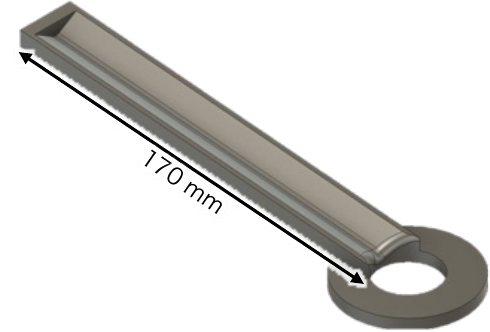}
            \subcaption{}
            \label{fig:jig:plumbing:groove}
        \end{minipage}
    \end{minipage} 
    \caption{Flow-plumbing jig receives liquid from end of tube in long groove then holds it in beaker. Servo motor can rotate groove to change flow of liquid. (a) Overview of jig. (b) Design of groove. (c) Design of funnel with gears.}
    \label{fig:jig:plumbing}
\end{figure*}

\subsubsection{Flow-plumbing Jig}
The flow-plumbing jig is designed to control the flow path of synthesized liquids in a chemical-synthesis system, enabling safe robotic sampling.
As illustrated in Fig. \ref{fig:jig:plumbing}, the jig consists of a funnel equipped with grooves and gears.
The groove that collects liquid from the tube is rotated to switch the flow path.

A gear mounted on the side of the funnel meshes with another gear connected to a servo motor, enabling the funnel to rotate and redirect the synthesized liquid.
The dual-gear configuration is intentionally designed to prevent synthesized liquid from contacting the servo motor.
When using the mobile manipulator or motion-demonstration interface to sample the liquid, a bottle is placed at the end of the tube, and the flow path is adjusted to guide the liquid into the bottle.

The FSM controlling the flow-plumbing jig has two states: ``sampling'' and ``disposal''.
Each control signal toggles between these states.
On the basis of the current state, the jig modifies the flow path accordingly to support proper liquid sampling or disposal.

\begin{figure*}[t]
    \centering
    \begin{minipage}[b]{0.49\linewidth}
        \centering
        \includegraphics[width=\hsize]{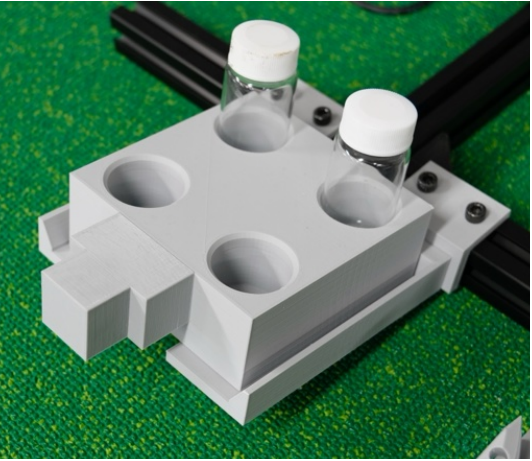}
        \subcaption{Overview}
        \label{fig:jig:bottle_holder:view}
    \end{minipage}
    \begin{minipage}[b]{0.49\linewidth}
        \centering
        \includegraphics[width=\hsize]{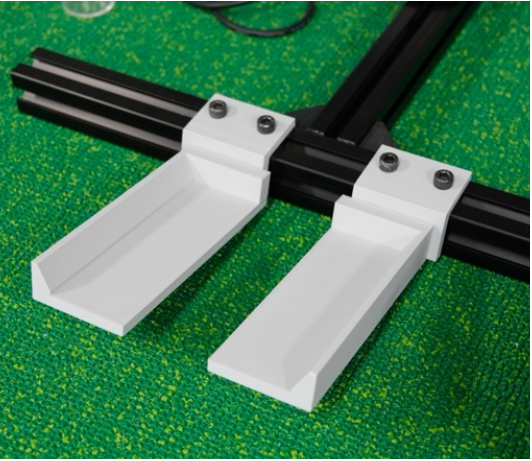}
        \subcaption{Bottle case stand}
        \label{fig:jig:bottle_holder:mounter}
    \end{minipage} \\
    \begin{minipage}[b]{0.3\linewidth}
        \centering
        \includegraphics[width=\hsize]{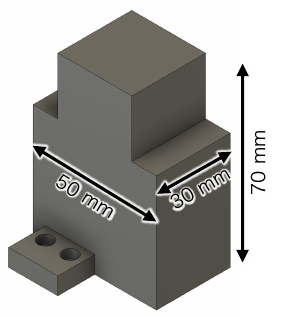}
        \subcaption{Handle}
        \label{fig:jig:bottle_holder:design}
    \end{minipage}
    \begin{minipage}[b]{0.6\linewidth}
        \centering
        \includegraphics[width=\hsize]{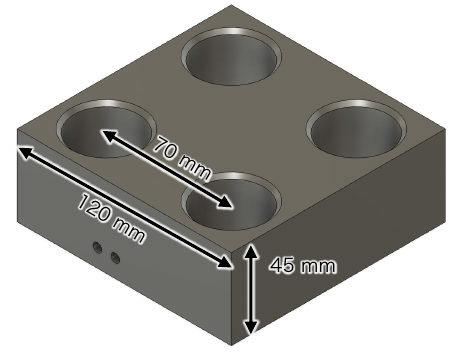}
        \subcaption{Bottle case}
        \label{fig:jig:bottle_holder:design}
    \end{minipage}
    \caption{Bottle-holder jig consists of the case that can store four bottles, with enough space between them for the mobile manipulator to pick up each bottle safely.}
    \label{fig:jig:bottle_holder}
\end{figure*}

\subsubsection{Bottle-holder Jig}
To enable the mobile manipulator or motion-demonstration interface to transport multiple bottles simultaneously, we developed a bottle holder capable of carrying four bottles at once, as shown in Fig. \ref{fig:jig:bottle_holder}.
This jig consists of a case with circular cutouts to hold the bottles and a handle for gripping by either the mobile manipulator or the motion-demonstration interface.
By grasping the handle, both the mobile manipulator's gripper and motion-demonstration interface can easily carry multiple bottles.
We also developed a sloped stand to assist in placing the case securely.

\subsection{Motion Demonstration and Execution}
This section describes the integration of the mobile manipulator, motion-demonstration interface, and experimental automation jigs within our chemical-experiment-automation system, detailing the method for automating experiments on the basis of chemists' motion demonstrations.
Demonstration data comprise the interface's positional information $\mathbf p^\mathrm{demo}_n$ and simultaneous state of the experimental-automation jig $\mathbf s^\mathrm{demo}_n$.
These collected data, consisting of $N$ data sequential end-effector points, are denoted as $\mathcal D=\{(\mathbf p^\mathrm{demo}_n, \mathbf s^\mathrm{demo}_n)\}^N_{n=1}$.
The robot needs to move to specific locations to reproduce each demonstration data, and these locations are assumed to be known in advance.

To reproduce the demonstrated motions, we employed proportional control (P-control) to compute the robot's end-effector velocity based on the positional error between the current and target pose.
While robot motion can be influenced by differences in payload mass, the controller does not consider these effects in our system, as the tasks are performed at low speeds and involve only lightweight items.
This control method was adopted considering its ease of use, real-time responsiveness, and stability when reproducing the demonstrated motions on the robot.
The mobile manipulator is controlled to target each data point in the collected demonstration dataset $\mathcal D$.
Given the current mobile manipulator position $\mathbf p_t^\mathrm{current}$ and current state of the experimental-automation jigs $\mathbf s_n^\mathrm{current}$, the robot-control command $\mathbf a_t^p$ and jig-control command $\mathbf a_t^s$ are calculated as
\begin{align}
 \mathbf{a}_t^p &= \alpha(\mathbf{p}^\mathrm{demo}_n - \mathbf{p}^\mathrm{current}_t), \\
 \mathbf{a}_t^s &= \mathrm{diff}(\mathbf{s}^\mathrm{demo}_n, \mathbf{s}^\mathrm{current}_t).
\end{align}
The robot control command $\mathbf a_t^p$ is determined from end-effector velocity and computed through proportional control (P-control), where $\alpha$ is the gain factor.
The jig control command $\mathbf a_t^s$ is determined by identifying state differences using the difference function $\mathrm{diff}(\cdot, \cdot)$.
The difference function $\text{diff}(\cdot, \cdot)$ compares the current state of each jig's FSM with the corresponding target state in the demonstration.
For each device, if a mismatch is detected, a predefined control command is triggered to transition the jig to the next state in its FSM.

The target positions for the manipulator and the desired states for the jigs are updated in a coordinated manner to synchronize their operations.
Specifically, when the robot's current pose is within an error threshold $\epsilon$ of the target pose, and the jig state matches the demonstrated state, the system increments the target index $n$ to proceed to the next step. 
This condition is evaluated in a control loop running at 20 (Hz), allowing the system to update the target index based on the latest tracking information from both the robot and the jigs.
This condition-based update mechanism ensures accurate and sequential reproduction of the demonstrated motions.
The threshold $\epsilon$ for updating the target position of the P-control is set to $1.0\times 10^{-3}$ (m).

The UR5e manipulator is controlled via Cartesian end-effector velocity commands provided by URScript. 
Specifically, the velocity commands obtained from P control are sent directly to the UR5e manipulator using the URScript ``speedl'' function. This approach is based purely on kinematic models and does not require inverse dynamics, torque level control, or joint inertia modeling.
Although no explicit collision avoidance or force-based safety constraints were implemented in this version of the system,  the UR5e has built-in protective stop features based on joint-torque monitoring, which were left enabled during all experiments to prevent damage in case of unexpected contact. 

The MiR200 mobile robot uses its built-in LiDAR sensor to generate a 2D map of the environment.
This map is then used for navigation during execution, allowing the robot to move to the predefined workplace autonomously. 
Our system assumes that the workspace configuration and task locations are known in advance, and that the MiR200 can reliably reach these specified positions.
To mitigate potential localization errors from MiR200's navigation, we incorporate a motion-capture system to estimate the precise position of the UR5e manipulator mounted on the mobile base.
The position information obtained from motion capture is then used to control the UR5e manipulator during task execution.
This approach allows us to compensate for base-level drift and ensures precise end-effector positioning.

\begin{figure*}[t]
    \centering
    \includegraphics[width=0.8\linewidth]{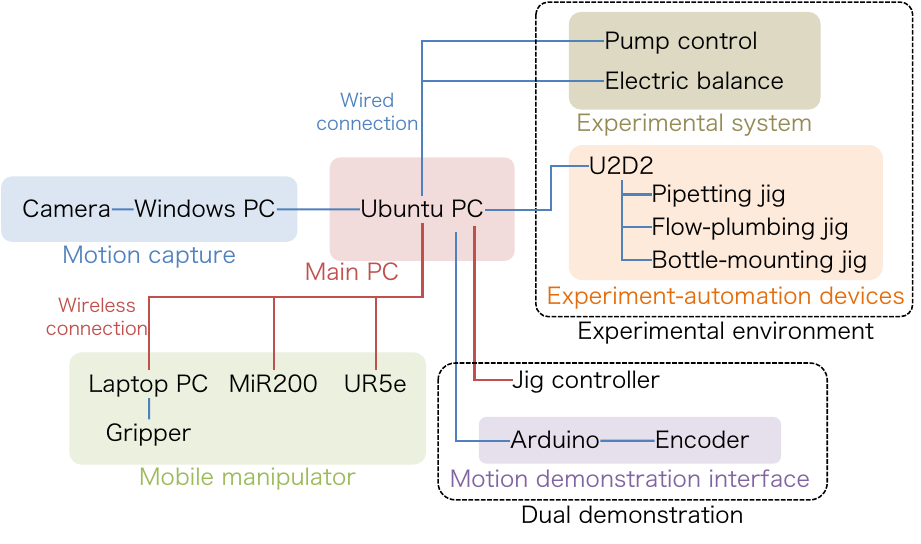}
    \caption{Overview of chemical-experiment-automation system}
    \label{fig:system:integration:overview}
\end{figure*}

\subsection{Integration of Robot Automation System}
Figure \ref{fig:system:integration:overview} provides an overview of the integrated experimental automation system, which combines the mobile manipulator, motion demonstration interface, and experimental automation jigs.
The system comprises seven main components: Main PC, motion capture system, experiment system, experimental automation jigs, mobile manipulator, motion demonstration interface, and jig controller.
The devices that our system consists of are summarized in Table \ref{tab:hardware_specs}.

All jigs in the experimental automation system are connected to a PC running Ubuntu 22.04 and are integrated via ROS Noetic. For experimental purposes, additional jigs such as a pump and an electronic balance were also programmed to communicate using Python and made compatible with ROS.

Chemists construct the experimental setup using both the experimental devices and the robotic automation jigs. To teach robotic actions, chemists use the motion demonstration interface and jig controller to input the necessary manipulations and control commands for the experiment.

\begin{table*}[t]
  \centering
  \caption{Hardware specifications and interfacing modes}
  \label{tab:hardware_specs}
  \begin{tabular}{p{2.5cm} p{2.5cm} p{4cm} p{4.0cm}}
    \toprule
    \textbf{Component} & \textbf{Device} & \textbf{Key Specs} & \textbf{Interface Mode} \\
    \midrule
    Mobile robot & MiR200 & Diff-drive, 200 (kg) payload, LiDAR nav. & Wi-Fi, ROS (onboard PC) \\
    Manipulator & UR5e & 6-DoF, 850 (mm) reach, $\pm$0.03 (mm) repeatability & URScript, ROS \\
    Gripper & RH-P12-RN & 0--106 (mm) opening, 2-finger parallel & Serial (Arduino), ROS \\
    Motion-demo interface & Custom & Mimics gripper shape, encoder-based sensing & Arduino Uno + encoder, ROS \\
    Motion capture & OptiTrack Flex13 & 120 (Hz), 3D marker-based tracking & Ethernet, Motive + ROS \\
    Experimental jigs & Dynamixel XM430-W350 & Servo (FSM-based), 3D-printed & U2D2 USB-serial, ROS \\
    \bottomrule
  \end{tabular}
\end{table*}

\section{Experiment}
We evaluated the effectiveness of the chemical-experiment-automation system by applying it to two types of chemical experiments.
Through these experiments, we verified that chemical tasks can be automated by a robot using dual demonstration provided via the motion-demonstration interface and jig controller.

\begin{figure*}[t]
    \centering
    \begin{minipage}[b]{0.6\linewidth}
        \centering
        \includegraphics[width=0.95\linewidth]{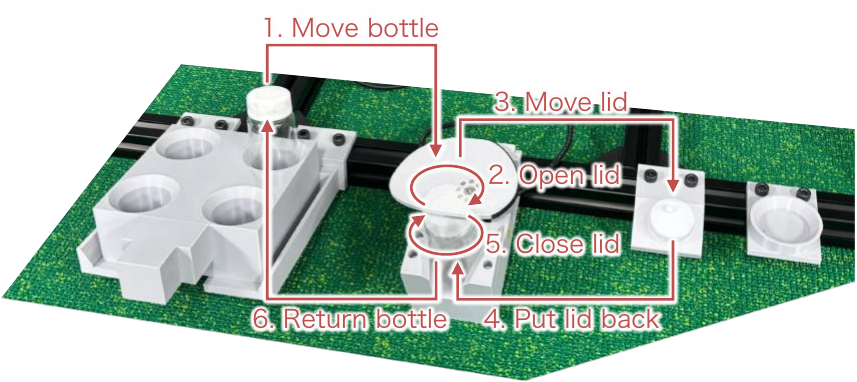}
        \subcaption{}
        \label{fig:exp1:env:bottle}
    \end{minipage}
    \begin{minipage}[b]{0.6\linewidth}
        \centering
        \includegraphics[width=0.95\linewidth]{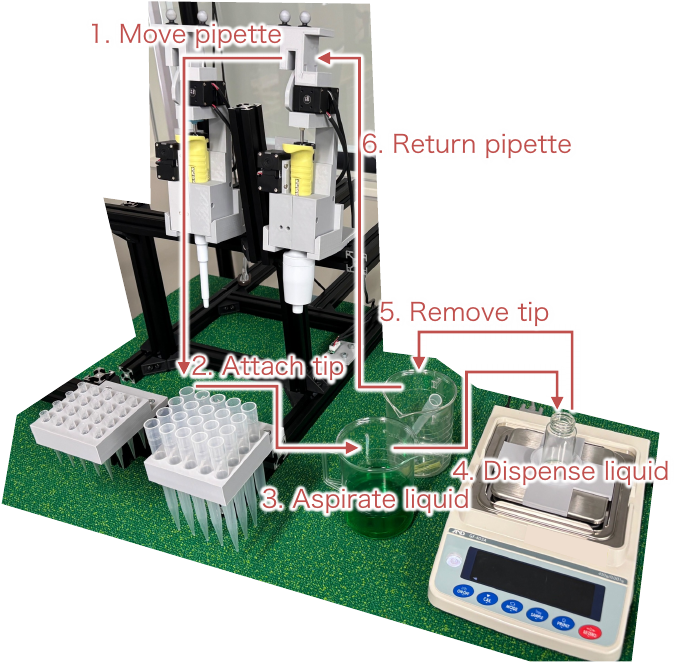}
        \subcaption{}
        \label{fig:exp1:env:pipette}
    \end{minipage} 
    \caption{Experimental environment for simple experimental tasks. (a) Bottle-manipulation task. (b) Pipetting task. Descriptions in figure show the steps of the demonstration.}
    \label{fig:exp1:env}
\end{figure*}

\subsection{Simple-task-automation Experiment}
\subsubsection{Settings}
To evaluate the reproduction performance of the system, we constructed two experimental environments focused on bottle and pipette manipulation-referred to as the bottle-manipulation task and pipetting task, respectively (Fig. \ref{fig:exp1:env}).
Figure \ref{fig:exp1:env:bottle} shows the experimental setup for the bottle-manipulation task developed using the bottle-holding and bottle-mounting jig.
In this task, the bottle is moved from the bottle case to the bottle mounter, the cap is opened and closed, and the bottle is returned to the case.
Figure \ref{fig:exp1:env:pipette} shows the pipetting-task setup constructed using the pipetting jig and electronic balance.
This task involves gripping the pipetting jig, attaching a tip, manipulating liquid, detaching the tip, and returning the pipetting jig.
In this setup, a chemist demonstrates using the pipette to transfer liquid into a bottle.
The chemist holds both the motion-demonstration interface and jig controller and uses the jig controller to trigger the recording of motion and send control commands to the jigs.
The gain $\alpha$ was empirically tuned through pilot tests to achieve smooth and stable tracking performance.
In our experiments, $\alpha$ was fixed at 4.0.

To evaluate the effectiveness of the system, the chemist conducted a single demonstration for each task, and the mobile manipulator executed each task ten times.
The system was assessed on the basis of three criteria: task-success rate, task-execution time, and motion reproducibility.
Task-execution time indicates the time taken from the start to the end of a manipulation motion.
During task execution, the mobile manipulator navigated to the task environment using the mobile base and replicated the demonstrated motions.
To evaluate the reproducibility of the mobile manipulator motions, we calculated the difference between its motions and the chemist's demonstrations using the Hausdorff distance.

\begin{table*}[t!]
    \centering
    \caption{Success rates, demonstration time, and task-execution times for bottle-manipulation and pipetting tasks. The execution time indicates the time it takes for the manipulator to reproduce the demonstration motion. The numbers in brackets indicate the standard deviation for ten trials.}
    \label{table:exp1:motion:eval}
    \begin{tabular}{c|cc}
        \toprule
                                 & Bottle manipulation & Pipetting \\
        \midrule
        Task-success rate        & 100 (\%)            & 100 (\%) \\
        Demonstration time (s)   & 57.4                & 41.2 \\
        Task-execution time (s)  & 161.3 (0.34)        & 115.6 (0.78) \\
        \bottomrule
    \end{tabular}
\end{table*}
\begin{table*}[t]
    \centering
    \caption{Trial-wise comparison of motion-reproduction accuracy between chemist's demonstration and robotic executions for both tasks. Values show the Hausdorff distances for position, orientation, and gripper width across 10 trials. To quantify variability, the mean and standard deviation are included in the last row.}
    \label{table:exp1:motion:error}
    \begin{tabular}{r|p{1.0cm} p{2.0cm} p{2.0cm} |p{1.0cm} p{2.0cm} p{2.0cm}}
        \toprule
              & \multicolumn{3}{c}{Bottle manipulation} & \multicolumn{3}{|c}{Pipetting} \\
        \midrule
        Trial & Position (mm) & Orientation (rad)       & Gripper width (mm) & Position (mm) & Orientation (rad)     & Gripper width (mm)  \\
        \midrule
        1st   & 9.78          & 6.20$\times 10^{-2}$  & 1.12 & 13.2          & 7.73$\times 10^{-2}$  & 1.33   \\
        2nd   & 8.67          & 6.40$\times 10^{-2}$  & 1.09 & 14.7          & 8.87$\times 10^{-2}$  & 1.33   \\
        3rd   & 9.57          & 6.49$\times 10^{-2}$  & 1.09 & 13.7          & 8.11$\times 10^{-2}$  & 1.33   \\
        4th   & 10.2          & 6.62$\times 10^{-2}$  & 1.09 & 12.9          & 7.08$\times 10^{-2}$  & 1.10   \\
        5th   & 8.50          & 6.41$\times 10^{-2}$  & 1.09 & 13.5          & 7.53$\times 10^{-2}$  & 1.33   \\
        6th   & 8.85          & 6.73$\times 10^{-2}$  & 1.04 & 13.5          & 7.63$\times 10^{-2}$  & 1.41   \\
        7th   & 9.23          & 6.45$\times 10^{-2}$  & 1.12 & 13.3          & 7.39$\times 10^{-2}$  & 0.937  \\
        8th   & 9.69          & 6.09$\times 10^{-2}$  & 1.12 & 13.4          & 8.07$\times 10^{-2}$  & 1.33   \\
        9th   & 9.60          & 6.35$\times 10^{-2}$  & 1.09 & 13.2          & 7.67$\times 10^{-2}$  & 0.938  \\
        10th  & 11.2          & 6.08$\times 10^{-2}$  & 1.20 & 13.3          & 8.56$\times 10^{-2}$  & 1.15   \\
        \midrule
        Mean (STD) & 9.53 (0.757)  & 6.38$\times 10^{-2}$ (2.03 $\times 10^{-3}$) & 1.11 (0.0396) 
                   & 13.5 (0.450)  & 7.86$\times 10^{-2}$ (5.17 $\times 10^{-3}$) & 1.22 (0.166)   \\
        \bottomrule
    \end{tabular}
\end{table*}

We specifically compared the time-series data of the position and orientation of the motion-demonstration interface from the chemist's demonstration $\mathbf X=\{\mathbf x_i\}_{i=1}^{N_\mathrm{ref}}$ with the end-effector trajectory of the mobile manipulator during task execution $\mathbf Y = \{\mathbf y_i\}_{i=1}^{N_\mathrm{robot}}$.
The Hausdorff distance was computed separately for position and orientation.
The distance function $d_f(\cdot, \cdot)$ used in the calculations used Euclidean distance for positions and angular differences for orientations.
The Hausdorff distance is defined as
\begin{align}
 d_H(\mathbf X, \mathbf Y) &=
 \mathrm{max}\left[
 \sup_{\mathbf x_i \in \mathbf X} d(\mathbf x_i, \mathbf Y),
 \sup_{\mathbf y_i \in \mathbf Y} d(\mathbf y_i, \mathbf X)
 \right] \\
 d(\mathbf x_i, \mathbf Y) &= \inf_{\mathbf y_j \in \mathbf Y} d_f(\mathbf x_i,\mathbf y_j).
\end{align}
This distance is a metric that captures the maximum deviation between two trajectories, providing a robust measure of motion error.

\begin{figure*}[t!]
    \centering
    \begin{minipage}[b]{\linewidth}
        \centering
        \includegraphics[width=0.6\linewidth]{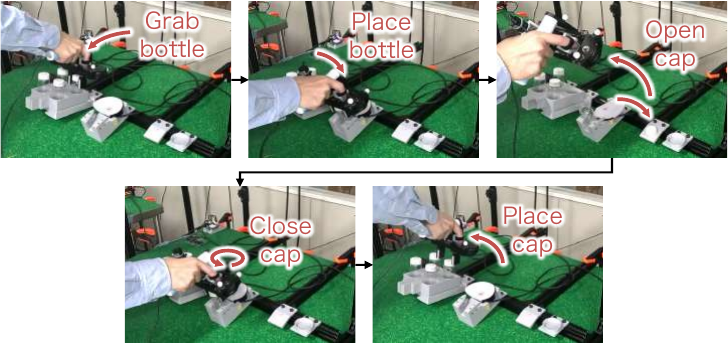}
        \subcaption{Motion demonstration}
        \label{fig:exp1:motion:bottle:demo}
    \end{minipage}\\
    \begin{minipage}[b]{\linewidth}
        \centering
        \includegraphics[width=0.6\linewidth]{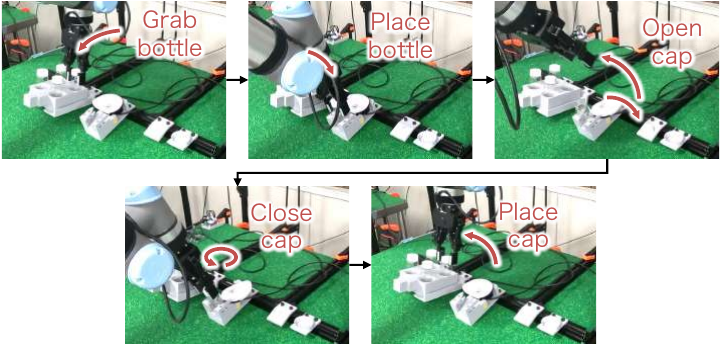}
        \subcaption{Robot execution}
        \label{fig:exp1:motion:bottle:exec}
    \end{minipage} 
    \caption{Demonstration motion and robot-executed motion in bottle-manipulation task}
    \label{fig:exp1:motion:bottle}
    \centering
    \begin{minipage}[b]{0.24\linewidth}
        \centering
        \includegraphics[width=0.95\hsize]{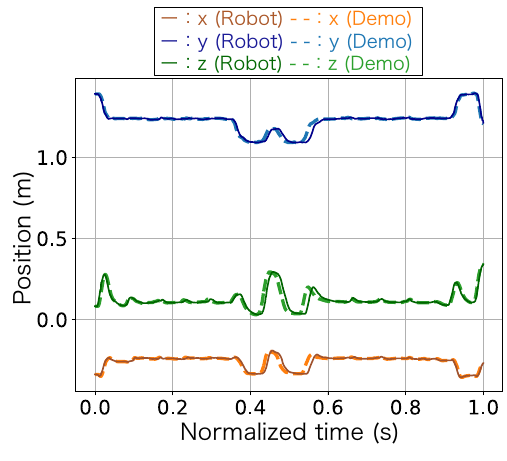}
        \subcaption{End-effector position}
        \label{fig:exp1:bottle:motion:seq:hand}
    \end{minipage} 
    \begin{minipage}[b]{0.24\linewidth}
        \centering
        \includegraphics[width=0.95\hsize]{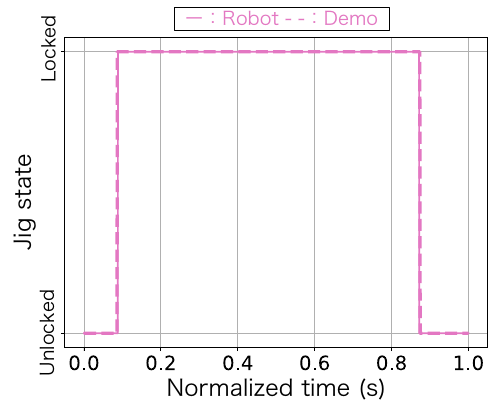}
        \subcaption{Jig operation}
        \label{fig:exp1:bottle:motion:seq:jig}
    \end{minipage} 
    \begin{minipage}[b]{0.24\linewidth}
        \centering
        \includegraphics[width=0.95\hsize]{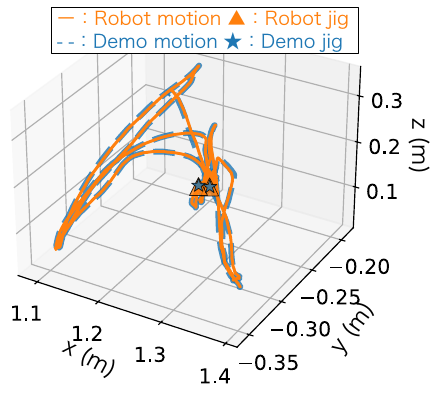}
        \subcaption{Motion trajectory}
        \label{fig:exp1:bottle:motion:seq:traj}
    \end{minipage}
    \begin{minipage}[b]{0.24\linewidth}
        \centering
        \includegraphics[width=0.95\hsize]{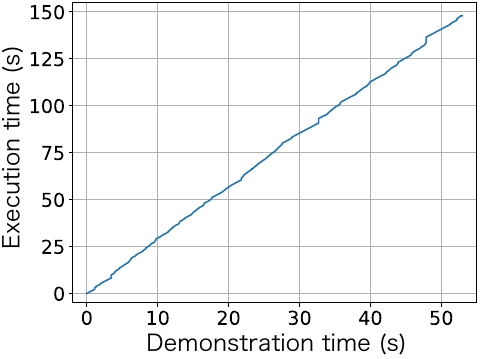}
        \subcaption{time-effector position}
        \label{fig:exp1:bottle:motion:seq:time}
    \end{minipage} 
    \caption{
    Visualization of demonstration and robotic execution in bottle-manipulation task.
    (a)  shows the transition of end-effector positions of the demonstration motion (dashed line) and the robot executed motion (solid line) with respect to normalized time on each axis (x, y, z).
    Here, normalized time indicates that the duration of each motion is normalized to the range from 0 to 1.
    (b) State transitions of bottle-mouting jig synchronized with end-effector positions with respect to normalized time.
    (c) 3D trajectory showing spatial points where jig operations were executed.
    These plots collectively illustrate the trajectory synchronization and phase alignment between the demonstrator and the robot.
    (d) shows the time-mapping plot obtained using dynamic time warping between the demonstration and execution motions.}
    \label{fig:exp1:bottle:motion:seq}
\end{figure*}

\begin{figure*}[!t]
    \centering
    \begin{minipage}[b]{0.49\linewidth}
        \centering
        \includegraphics[width=0.95\linewidth]{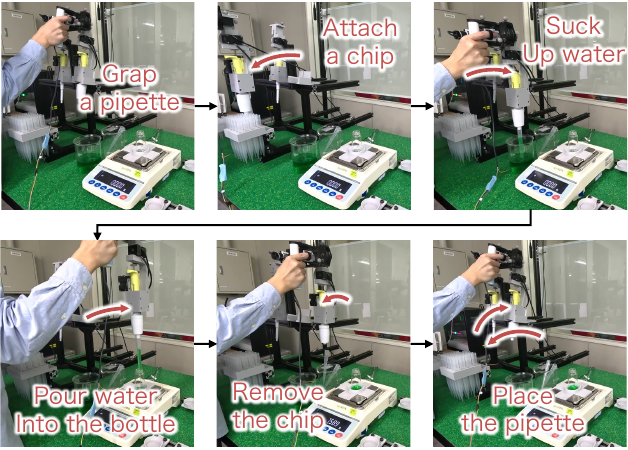}
        \subcaption{Motion demonstration}
        \label{fig:exp1:motion:pipetting:demo}
    \end{minipage} 
    \begin{minipage}[b]{0.49\linewidth}
        \centering
        \includegraphics[width=0.95\linewidth]{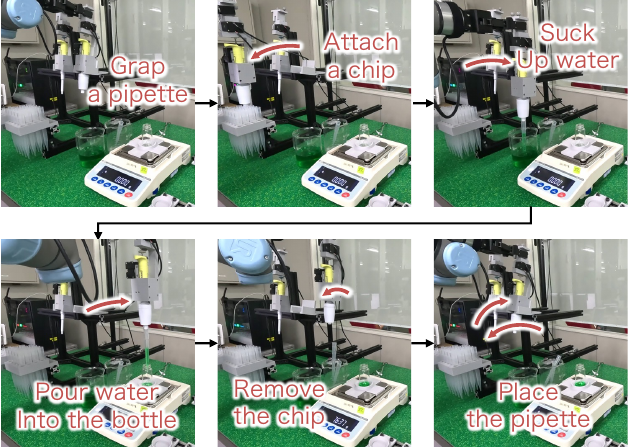}
        \subcaption{Robot execution}
        \label{fig:exp1:motion:pipetting:exec}
    \end{minipage} 
    \caption{Demonstration motion and robot-executed motion in pipetting task}
    \label{fig:exp1:motion:pipetting}
\end{figure*}
\begin{figure*}[t]
    \centering
    \begin{minipage}[b]{0.24\linewidth}
        \centering
        \includegraphics[width=0.95\hsize]{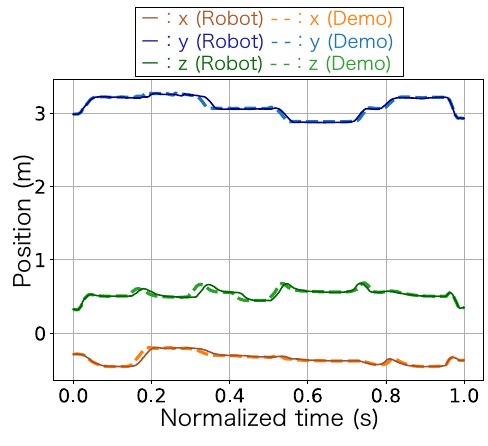}
        \subcaption{End-effector position}
        \label{fig:exp1:pipette:motion:seq:hand}
    \end{minipage} 
    \begin{minipage}[b]{0.24\linewidth}
        \centering
        \includegraphics[width=0.95\hsize]{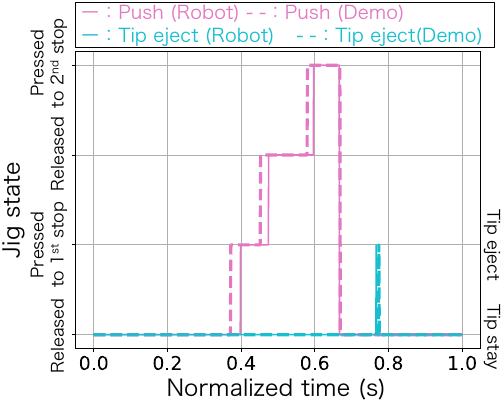}
        \subcaption{Jig operation}
        \label{fig:exp1:pipette:motion:seq:jig}
    \end{minipage}  
    \begin{minipage}[b]{0.24\linewidth}
        \centering
        \includegraphics[width=0.95\hsize]{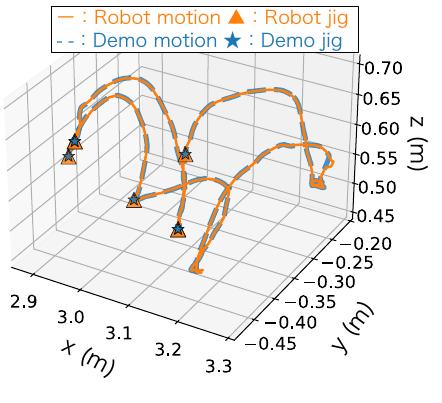}
        \subcaption{Motion trajectory}
        \label{fig:exp1:pipette:motion:seq:traj}
    \end{minipage}
    \begin{minipage}[b]{0.24\linewidth}
        \centering
        \includegraphics[width=0.95\hsize]{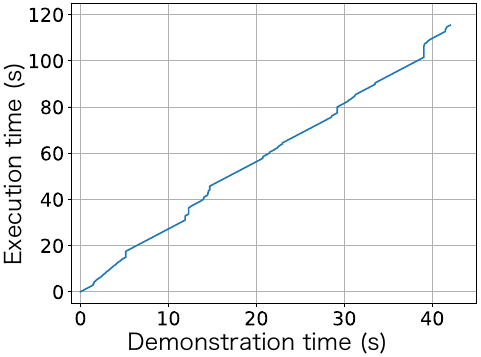}
        \subcaption{End-effector position}
        \label{fig:exp1:pipette:motion:seq:hand}
    \end{minipage} 
    \caption{
    Visualization of demonstration and robotic execution in pipetting task.
    (a) shows the transition of end-effector positions of the demonstration motion (dashed line) and the robot executed motion (solid line) with respect to normalized time on each axis (x, y, z).
    Here, normalized time indicates that the duration of each motion is normalized to the range from 0 to 1.
    (b) State transitions of pipetting jig synchronized with end-effector positions with respect to normalized time.
    (c) 3D trajectory showing spatial points where jig operations were executed.
    These plots collectively illustrate the trajectory synchronization and phase alignment between the demonstrator and the robot.
    (d) shows the time-mapping plot obtained using dynamic time warping between the demonstration and execution motions.
    }
    \label{fig:exp1:pipette:motion:seq}
\end{figure*}

\subsubsection{Results}
Table \ref{table:exp1:motion:eval} presents the task-success rates and execution times for the mobile manipulator and motion demonstration across the two evaluated tasks.
In all ten trials for each task, the mobile manipulator successfully completed the assigned task.

Regarding execution time which indicates the time it takes for the manipulator to reproduce the demonstration motion,
both tasks required approximately 2.8 times longer when executed by the mobile manipulator compared with the chemist demonstration.
Table \ref{table:exp1:motion:error} shows the Hausdorff distances between the motion demonstration and each of the ten robotic executions for each task.
The differences in position, orientation, and gripper width were all minimal, indicating high fidelity in motion reproduction.
The position error includes both control and measurement error.
However, since the OptiTrack Flex 13 system has a positional accuracy of $\pm$ 0.1 (mm), and the observed position errors exceed this range, we attribute the majority of the error to tracking limitations in the robot control.
These results indicate that the system can accurately replicate demonstrated actions using a mobile manipulator, enabling effective automation of experimental tasks.

Figures \ref{fig:exp1:motion:bottle} and \ref{fig:exp1:bottle:motion:seq} respectively show snapshots and detailed motion visualizations for the bottle-manipulation task during both demonstration and robotic execution.
The snapshots in Fig. \ref{fig:exp1:motion:bottle} show that the mobile manipulator accurately replicated the demonstrated motions.
Fig \ref{fig:exp1:bottle:motion:seq} (a) to (c) represent the end-effector trajectories and jig state transitions with respect to normalized time, where normalized time denotes that the duration of each motion is scaled from 0 to 1.
This duration time also indicates the time taken for the motion-demonstration interface and UR5e manipulator movements.
The results show that although the robot's motions were delayed behind the demonstration, jig operations were executed at the same spatial locations, confirming synchronization between robotic motion and jig operation.

\begin{figure*}[!t]
    \centering
    \begin{minipage}[b]{\linewidth}
        \centering
        \includegraphics[width=0.6\hsize]{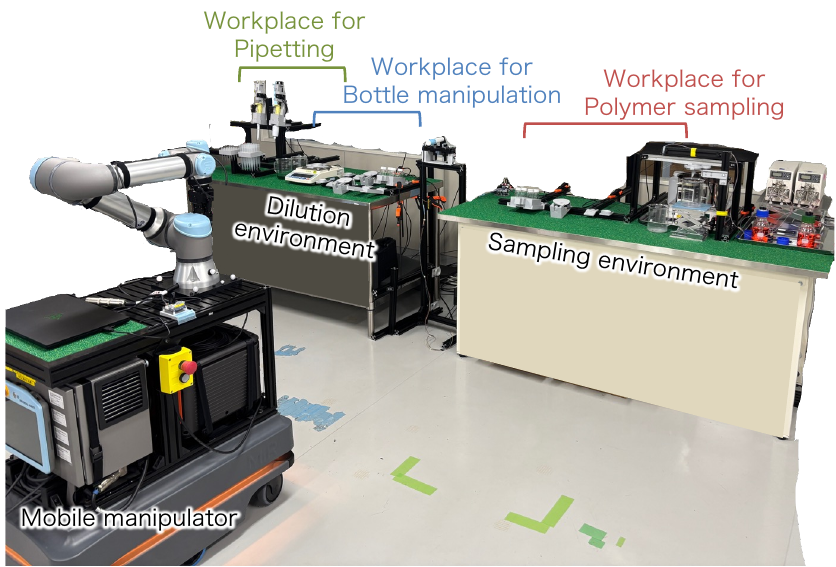}
        \subcaption{}
        \label{fig:exp2_experiment_env:env}
    \end{minipage} \\
    \begin{minipage}[b]{0.49\linewidth}
        \centering
        \includegraphics[width=0.8\hsize]{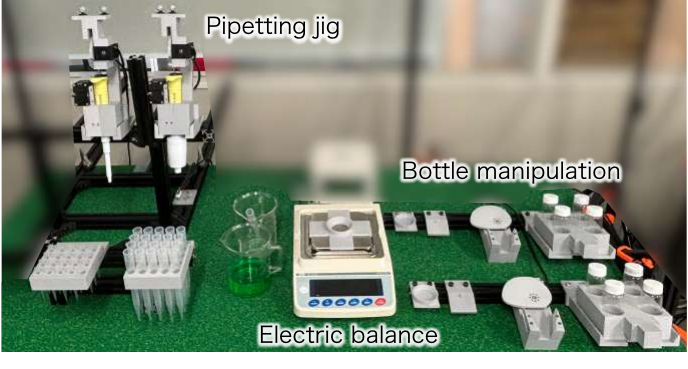}
        \subcaption{}
        \label{fig:exp2_experiment_env:kisyaku}
    \end{minipage}
    \begin{minipage}[b]{0.49\linewidth}
        \centering
        \includegraphics[width=0.8\hsize]{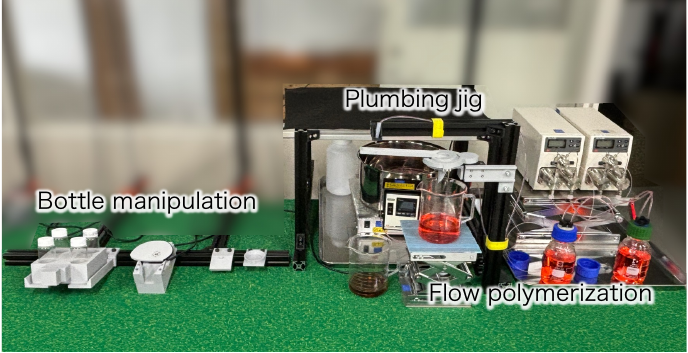}
        \subcaption{}
        \label{fig:exp2_experiment_env:synthesis}
    \end{minipage}
    \caption{Overview of experimental environment. (a) Layout of the experimental system with three designated workspaces: polymer sampling, bottle manipulation, and pipetting. (b) Environment for bottle manipulation and pipetting. (c) Environment for polymer sampling.}
    \label{fig:exp2_experiment_env}
\end{figure*}

Figure \ref{fig:exp1:bottle:motion:seq:time} presents the time-mapping plot obtained using dynamic time warping (DTW), which quantifies the temporal correspondence between demonstration and execution motions.
The nearly linear time-mapping relationship suggests that, irrespective of the specific motions in the bottle-manipulation task, the controller induced a consistent delay in execution relative to the demonstration.

Figures \ref{fig:exp1:motion:pipetting} and \ref{fig:exp1:pipette:motion:seq} present corresponding results for the pipetting task.
As with the bottle-manipulation task, the normalized-time plots show that robot motions are maintained in alignment in the timing and location of jig operations.
The DTW-based time-mapping again confirms that, irrespective of the temporal delay, the controller maintained the correct motion, ensuring successful task execution by the manipulator.

\subsection{Laboratory-automation Experiment}
To further evaluate the performance and practical applicability of the system, we applied it to a simulated environment for polymer synthesis.

\subsubsection{Settings}
In this experiment, we assumed that a typical polymer-synthesis task consists of three subtasks: polymer sampling, bottle manipulation, and dilution via pipetting.
On the basis of this assumption, we constructed a simulated laboratory environment, as shown in Fig. \ref{fig:exp2_experiment_env}.
With this setup, it was assumed that the synthesis and dilution steps were conducted at separate workstations.
The system consists of a sampling environment that includes a pump system and bottle-manipulation jigs for flow synthesis and a dilution environment that incorporates pipettes and bottle holders.
Both the demonstrator and mobile manipulator move between three designated work positions to complete the entire polymer-synthesis procedure.
At each position, the following tasks are performed: 1) bottle manipulation and mock-polymer sampling, 2) preparation of bottles for dilution, and 3) execution of the dilution process via pipette manipulation.

In the polymer-sampling task, one of the bottles is retrieved from a bottle case, and a mock-polymer liquid (colored water) is collected using the bottle from a pump.
During the bottle-preparation step for dilution, two bottles are arranged-one containing the mock polymer and the other intended for the diluted solution.
The dilution operation involves collecting 0.600 mL of the mock polymer using a small micropipette and 8.00 mL of diluent using a large micropipette.
The step-by-step procedures are outlined in Work Procedures \ref{procedure:sampling}, \ref{procedure:dilution:bottle}, and \ref{procedure:dilution:pipette}.
To evaluate the effectiveness of the system, the mobile manipulator was instructed using a single demonstration by the chemist then executed the experiment three times.
The evaluation criteria include task-success rate and the composition of the diluted mock-polymer solution obtained through the robotic executions.
The gain $\alpha$ of the robot controller is set to 4.0, as in the simple-task-automation experiment.

\begin{algorithm}[t]
    \SetAlgoLined
    \DontPrintSemicolon
    Mobile manipulator moves to workspace 1\;
    Retrieves a bottle from the bottle case.\;
    Opens the lid of the bottle.\;
    Uses the bottle to sample the polymer.\;
    Closes the lid of the bottle.\;
    Places the bottle containing the polymer back into the case.\;
    Lifts the bottle case.\;
    \caption{Polymer sampling task}
    \label{procedure:sampling}
\end{algorithm}

\begin{algorithm}[!t]
    \SetAlgoLined
    \DontPrintSemicolon
    Mobile manipulator moves to workspace 2\;
    Places the bottle case\;
    Removes the bottle containing the polymer from the case\;
    Opens the lid of the bottle containing the polymer\;
    Removes the dilution bottle from the case\;
    Opens the lid of the dilution bottle\;
    Places the dilution bottle on the digital balance\;
    \caption{Bottle manipulation for polymer dilution}
    \label{procedure:dilution:bottle}
\end{algorithm}

\begin{algorithm}[t]
    \SetAlgoLined
    \DontPrintSemicolon
    Mobile manipulator moves to workspace 3\;
    Grasps a small micropipette\;
    Attaches the tip to the small micropipette\;
    Puts the polymer into the dilution bottle by using the small micropipette\;
    Removes the tip from the small micropipette\;
    Places the small micropipette\;
    Grasps a large micropipette\;
    Attaches the tip to the large micropipette\;
    Puts the solution into the dilution bottle by using the large micropipette\;
    Removes the tip from the large micropipette\;
    Places the small micropipette\;
    Closes the lid on the bottle containing the polymer\;
    Closes the lid on the dilution bottle\;
    Puts the bottle containing the polymer into the case\;
    Puts the dilution bottle into the case\;
    \caption{Pipetting for polymer dilution}
    \label{procedure:dilution:pipette}
\end{algorithm}

\begin{table}[!t]
    \caption{Amount of liquid pipetted in three trials. Target volumes of polymer and solvent were 0.60 mL and 8.0 mL, respectively. The mean and standard deviation are included in the last row.}
    \label{table:exp2:pipette}
    \begin{tabular}{c|cc}
        \toprule
        Trial & Polymer (g) & Solvent (g)\\
        \midrule
        1st   & 0.585       & 7.615 \\
        2nd   & 0.590       & 7.700 \\
        3rd   & 0.585       & 7.700 \\
        \midrule
        Mean (STD) & 0.587 (2.36 $\times 10^{-3}$)  & 7.67 (4.01 $\times 10^{-2}$)  \\
        \bottomrule
    \end{tabular}
\end{table}

\begin{figure*}[t]
    \centering
    \includegraphics[width=0.95\hsize]{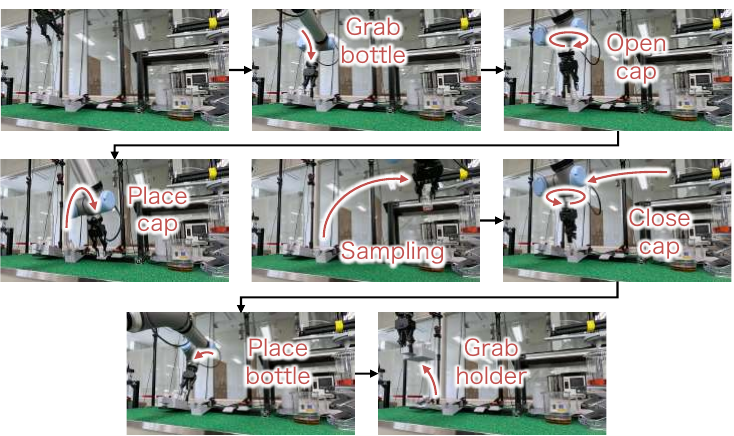}
    \caption{Robot-executed motion for sampling on workspace 1}
    \label{fig:robot_exp:sampling}
\end{figure*}
\begin{figure*}[t]
    \centering
    \includegraphics[width=0.95\hsize]{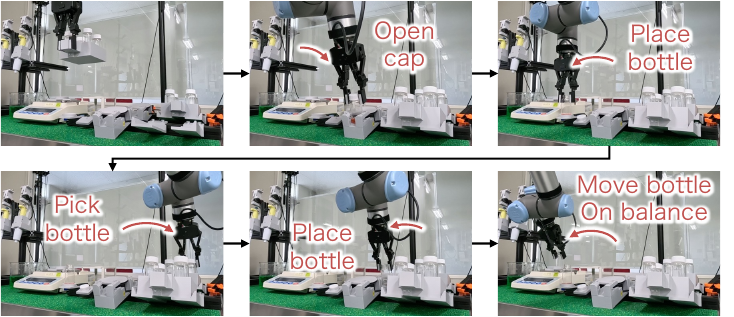}
    \caption{Robot-executed motion for bottle manipulation on workspace 2}
    \label{fig:robot_exp:bottle}
\end{figure*}
\begin{figure*}[t]
    \centering
    \includegraphics[width=0.95\hsize]{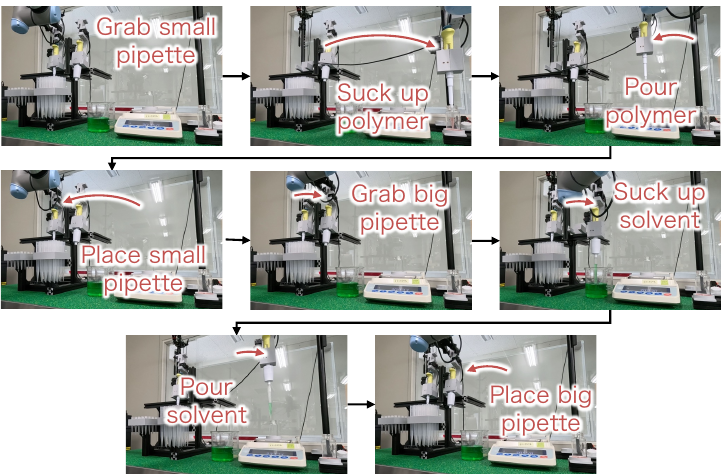}
    \caption{Robot-executed motion for pipetting on workspace 3}
    \label{fig:robot_exp:pipette}
\end{figure*}

\subsubsection{Results}
Using the system, the experiment was executed three times, and in all three trials, the mobile manipulator successfully completed the entire sequence of experimental actions.
Table \ref{table:exp2:pipette} shows the volumes of the two types of liquids transferred using two pipettes during the three trials.
The target volumes were 0.600 mL for the polymer and 8.00 mL for the solvent.
The measured results indicate that the mobile manipulator handled the liquids accurately in accordance with the specified quantities.
Figures \ref{fig:robot_exp:sampling}, \ref{fig:robot_exp:bottle}, and \ref{fig:robot_exp:pipette} illustrate the mobile manipulator's actions during polymer sampling, bottle manipulation for dilution, and pipette operations, respectively, as executed by the system in the simulated polymer-synthesis experiment.
These results confirm that the proposed concept enables the automation of chemical-experiment processes through robotic execution of chemist-taught demonstrations, effectively validating the practicality and precision of our system.

\section{Discussions}

\subsection{Limitation of Current System}
Although our chemical-experiment-automation system demonstrated promising results, it also has several limitations.
The system currently replicates demonstrated motions without adaptation; therefore, it cannot adjust its motions in response to changes in the positions of target objects or experimental equipment.
The performance of the system is also highly dependent on the accuracy of position estimation using the motion-capture system, and even small errors can significantly affect the task-success rate.
Specifically, the system assumes consistent positioning of experimental jigs between the demonstration and execution. 
Even slight misalignments in jig positions can lead to execution failures or degraded precision, as the robot lacks a mechanism for real-time correction.

To enhance robustness, future versions of the system could incorporate real-time feedback mechanisms such as vision-based jig localization or tactile and force sensing for contact-aware manipulation.
In particular, force-sensitive operations, such as pipette insertion, lid closing, or liquid handling, could benefit from compliance control using external force-torque sensors or the built-in force sensing capabilities of the UR5e manipulator.
Such sensing would help absorb minor positional errors and stabilize interactions with delicate or deformable objects.

Integrating these sensing modalities would allow the robot to adapt to minor variations in the environment, improving both reliability and generalizability in practical laboratory settings.
To select robot actions based on perception from tactile sensors, camera images, and force measurements, it is necessary to apply policy learning methods such as imitation learning \cite{osa2018} and reinforcement learning \cite{hazem2025, benhazem2025a}.

The system also requires complete and consistent motion demonstrations.
Any minor modification to the task requires re-demonstration of the entire sequence.
To address this limitation, techniques that automatically segment demonstrations into semantically meaningful sub-actions \cite{lea2017, ding2023b} and systems that support motion scheduling \cite{yoshikawa2023c} could enable partial re-teaching or re-composition of robotic actions from previously collected demonstrations.

\subsection{Application and Extension}
This study demonstrated the feasibility of our system only in a polymer-synthesis setting.
However, our system has the potential to be applied to a wide range of domains because of its modular architecture.
For example, by modifying the handling motion of the pipetting tool and bottles, the system could be deployed in automated workflows for bioassay analysis, pharmaceutical liquid dispensing, or contamination inspection.
Exploring such applications will be a focus of our future work.

While our current system is designed for a single-arm mobile manipulator, extending the dual demonstration concept to multi-agent robotic systems (e.g., bimanual manipulators or multi-robot platforms) could significantly improve efficiency in chemical experiment automation.
For instance, one agent could prepare samples while another performs pipetting or measurement, thereby enabling parallelized workflows across multiple workstations.
However, such an extension is not straightforward.
Multi-agent coordination requires not only the replication of individual demonstrated motions but also the modeling of inter-agent dependencies.
In chemical experiments, the sequence and timing of motions must often be precisely aligned.
Thus, agents must choose motions with consideration for others' states and intentions.
Incorporating such coordination into the learning or execution process requires coordinated mechanisms \cite{le2017, seo2025a}, which go beyond the current dual-demonstration setup.
Addressing these challenges is a promising direction for future research.

\subsection{Jig Controller Design}
In the current system, jig controllers are implemented using FSMs, which are well-suited for handling discrete state transitions.
This design choice suits the human demonstration, which issues discrete commands to the jigs via a simple controller.
While replacing the FSMs with continuous controllers such as those based on reinforcement learning or trajectory optimization could enable more sophisticated and precise control, it would also require the demonstrator to specify control inputs in a more continuous and potentially complex manner. 
This introduces a trade-off between control flexibility and ease of demonstration.
Therefore, the integration of continuous controllers must be carefully designed to maintain the intuitiveness and accessibility of the demonstration interface.

\subsection{Motion Control and Stability}
Although our system adopts a simple proportional controller for end-effector velocity control, we observed stable convergence behavior in all evaluated tasks.
This is due to the relatively low robot movement speed in chemical manipulation tasks, which reduces the influence of dynamic effects and minimizes oscillatory behavior.
However, the stability of this open-loop proportional control may degrade under more dynamic or contact-sensitive conditions, such as high-speed pick-and-place operations or tasks requiring precise force regulation.
In such scenarios, factors like unmodeled dynamics, payload variation, and interaction forces could lead to divergence or unsafe motions if not adequately compensated.
Incorporating adaptive gain tuning or switching to impedance control frameworks could further enhance stability and safety, especially in tasks involving force-sensitive operations or environmental uncertainties.
Adopting adaptive or model-informed control strategies \cite{benhazem2025} for this system may ensure robustness against environmental and payload variations while retaining the intuitive programming benefits of the dual demonstration approach.

\section{Conclusions}
We proposed a dual-demonstration concept for automating chemical experiments, in which chemists can simultaneously demonstrate robotic motions and jig operations using an intuitive interface.
To validate this concept, we developed a chemical-experiment-automation system that includes a motion-demonstration interface, a set of experimental jigs, and a mobile manipulator integrated via ROS.

Our main contributions are as follows:
(i) the introduction of a novel dual-demonstration paradigm enabling coordinated robot and jig programming through human demonstrations,
(ii) the development of hardware and software infrastructure to support this paradigm in real experimental environments, and
(iii) the successful execution of polymer synthesis tasks with high reproducibility and robustness, confirming the feasibility of the system.

Experimental results demonstrated that the robot could accurately reproduce human-taught operations with minimal deviation, maintaining 100\% task success rates in both simple and compound workflows.
These findings confirm that dual demonstration is a viable alternative to traditional robot programming, particularly in laboratory environments where flexibility and adaptability are required.
In terms of practical impact, the system enables chemists without robotics expertise to automate complex laboratory tasks through intuitive interaction, significantly reducing the programming burden and opening the door for broader adoption of robotic lab automation.

As future work, we plan to address current limitations related to environmental variability and sensor dependency. In particular, we aim to integrate visual and tactile feedback to improve robustness and enable adaptive execution.

\section*{Acknowledgements}
This paper is based on results obtained from a project, JPNP14004, subsidized by the New Energy and Industrial Technology Development Organization (NEDO).



\bibliography{reference}

\end{document}